%% file: paper.tex
\documentclass{article}

\usepackage{microtype}
\usepackage{graphicx}
\usepackage{booktabs} 

\usepackage{hyperref}

\usepackage[accepted]{icml2023}

\usepackage{amsmath}
\usepackage{amssymb}
\usepackage{mathtools}
\usepackage{amsthm}

\usepackage[capitalize,noabbrev]{cleveref}

\usepackage{subcaption} 
\usepackage{comment}
\usepackage{url}
\usepackage{caption}
\usepackage{subcaption}
\usepackage{wrapfig}
\usepackage{floatflt}
\usepackage{multirow}
\usepackage{tikz}
\usepackage{rotating}

\theoremstyle{plain}

\theoremstyle{definition}

\theoremstyle{remark}

\usepackage[textsize=tiny]{todonotes}

\icmltitlerunning{CLUTR: Curriculum Learning via Unsupervised Task Representation Learning}

\begin{document}

\twocolumn[
\icmltitle{CLUTR: Curriculum Learning via Unsupervised Task Representation Learning}





\begin{icmlauthorlist}

\icmlauthor{Abdus Salam Azad}{berkeley}
\icmlauthor{Izzeddin Gur}{google}
\icmlauthor{Jasper Emhoff}{berkeley}
\icmlauthor{Nathaniel Alexis}{berkeley}
\icmlauthor{Aleksandra Faust}{google}
\icmlauthor{Pieter Abbeel}{berkeley}
\icmlauthor{Ion Stoica}{berkeley}
\end{icmlauthorlist}

\icmlaffiliation{berkeley}{University of California, Berkeley}
\icmlaffiliation{google}{Google Research}

\icmlcorrespondingauthor{Abdus Salam Azad}{salam\_azad@berkeley.edu}

\icmlkeywords{Machine Learning, Reinforcement Learning}

\vskip 0.3in
]



\printAffiliationsAndNotice{}  

\begin{abstract}
\input{sections/abstract.tex}
\end{abstract}

\section{Introduction}
\input{sections/intro.tex}

\section{Related Work}\label{sec:related}\input{sections/related_works.tex}
\section{Background}\input{sections/background.tex}
\section{Curriculum Learning via Unsupervised Task Representation Learning}
\input{sections/clutr_method.tex}

\section{Experiments}
\input{sections/experiments.tex}
\section{Conclusion: Limitations and Future Work}
\input{sections/conclusions.tex}
\newpage

\bibliography{paper}
\bibliographystyle{icml2023}

\newpage

\section*{Acknowledgments}

This project is a collaboration under the Google-BAIR (Berkeley Artifcial Intelligence Research) Commons program. It was supported in part by NSF CISE Expeditions Award CCF-1730628, in  addition to gifts from Astronomer, Google, IBM, Intel, Lacework, Microsoft, Mohamed Bin Zayed University of Artificial Intelligence, Nexla, Samsung SDS, Uber, and VMware. We gratefully acknowledge Natasha Jaques for their valuable feedback. We also acknowledge Raymond Chong, Adrian Liu, and Sarah Bhaskaran for the valuable discussions.

\section*{Ethic Statement}
Unsupervised Environment Design can be applied to many real-world applications and shares similar ethical concerns and considerations with other Artificially Intelligent(AI) systems. For example, AI systems can cause more unemployment or be used for reasons/applications that have a negative societal impact, for which responsible usage of such AI systems must be promoted and established. During our research, all the experiments were done in simulation and no human or living subjects were used. 

\section*{Reproducibility}
Our code, saved checkpoints, and training data are available at \url{https://github.com/clutr/clutr}


\newpage
\appendix
\onecolumn
\input{sections/appendices.tex}

\end{document}

%% file: sections/abstract.tex
Reinforcement Learning (RL) algorithms are often known for sample inefficiency and difficult generalization. Recently, Unsupervised Environment Design (UED) emerged as a new paradigm for zero-shot generalization by simultaneously learning a task distribution and agent policies on the generated tasks. This is a non-stationary process where the task distribution evolves along with agent policies; creating an instability over time. While past works demonstrated the potential of such approaches, sampling effectively from the task space remains an open challenge, bottlenecking these approaches. To this end, we introduce CLUTR: a novel unsupervised curriculum learning algorithm that decouples task representation and curriculum learning into a two-stage optimization. It first trains a recurrent variational autoencoder on randomly generated tasks to learn a latent task manifold. Next, a teacher agent creates a curriculum by maximizing a minimax REGRET-based objective on a set of latent tasks sampled from this manifold. Using the fixed-pretrained
task manifold, we show that CLUTR successfully overcomes the non-stationarity problem and improves stability. Our experimental results show CLUTR outperforms PAIRED, a principled and popular UED method, in the challenging CarRacing and navigation environments: achieving 10.6X and 45\% improvement in zero-shot generalization, respectively. CLUTR also performs comparably to the non-UED state-of-the-art for CarRacing, while requiring 500X fewer environment interactions. 

%% file: sections/intro.tex
Deep Reinforcement Learning (RL) has shown exciting progress in the past decade in many challenging domains including Atari~\cite{atari}, Dota~\cite{dota}, Go~\cite{go}. However, deep RL is also known for its sample inefficiency and difficult generalization---performing poorly on unseen tasks or failing altogether with the slightest change~\cite{generalization0, generalization1, generalization2}. While, Curriculum Learning (CL) algorithms have shown to improve RL sample efficiency by adapting the training task distribution, i.e., the curriculum~\cite{cl2,cl1}, recently a class of Unsupervised CL algorithms, called Unsupervised Environment Design (UED)~\cite{paired, dcd} has shown promising zero-shot generalization by automatically generating the training tasks and adapting the curriculum simultaneously.

UED algorithms employ a teacher that generates training tasks by sampling the free parameters of the environment (e.g.,  the start, goal, and obstacle locations for a navigation task) and can either be adaptive or random. Contemporary adaptive UED teachers, i.e., PAIRED ~\cite{paired} and REPAIRED~\cite{dcd}, are implemented as RL agents with the free task parameters as their action space. The teacher agent aims at generating tasks that maximize the student agent's regret, defined as the performance gap between the student agent and an optimal policy. Inspite of promising zero-shot generalization, adaptive teacher UEDs are still sample inefficient.

This sample inefficiency is attributed primarily to the difficulty of training a regret based RL teacher~\cite{accel}. First, the teacher receives a sparse reward only after specifying the full parameterization of a task; leading to a long-horizon credit assignment problem. Additionally, the teacher agent faces a combinatorial explosion problem if the parameter space is permutation invariant---e.g., for a navigation task, a set of obstacles corresponds to factorially different permutations of the parameters\footnote{Consider a 13x13 grid for a navigation task, where the locations are numbered from 1 to 169. Also consider a wall made of four obstacles spanning the locations: \{21, 22, 23, 24\}. This wall can be represented using any permutation of this set, e.g., \{22, 24, 23, 21\}, \{23, 21, 24, 22\}, resulting in a combinatorial explosion.}. Most importantly, the teacher needs to simultaneously learn a task manifold--from scratch--to generate training tasks and navigate this manifold to induce an efficient curriculum. However, the teacher learns this task manifold implicitly based on the student regret and  as the student is continuously co-learning with the teacher, the task manifold also keeps evolving over time. Hence, the simultaneous learning of task manifold and curriculum results in an instability over time and makes it a difficult learning problem. 


To address the above-mentioned challenges, we present Curriculum Learning via Unsupervised Task Representation Learning (CLUTR). At the core of CLUTR, lies a hierarchical graphical model that decouples task representation learning from curriculum learning. We develop a variational approximation to the UED problem and employ a Recurrent Variational AutoEncoder (VAE) to learn a latent task manifold, which is pretrained unsupervised. Unlike contemporary adaptive-teachers, which builds the tasks from scratch one parameter at a time, the CLUTR teacher generates tasks in a single timestep by sampling points from the latent task manifold and uses the generative model to translate them into  complete tasks. The CLUTR teacher learns the curriculum by navigating the pretrained and fixed task manifold via maximizing regret. By utilizing a pretrained latent task-manifold, the CLUTR teacher can train as a contextual bandit -- overcoming the long-horizon credit assignment problem -- and create a curriculum much more efficiently -- improving stability at no cost to its effectiveness. Finally, by carefully introducing bias to the training corpus (such as sorting each parameter vector), CLUTR solves the combinatorial explosion problem of parameter space without using any costly environment interactions.

While CLUTR can be integrated with any adaptive teacher UEDs, we implement CLUTR on top of PAIRED---one of the most principled and popular UEDs. Our experimental results show that CLUTR outperforms PAIRED, both in terms of generalization and sample efficiency, in the challenging pixel-based continuous CarRacing and partially observable discrete navigation tasks. For CarRacing, CLUTR achieves 10.6X higher zero-shot generalization on the F1 benchmark~\cite{dcd} modeled on 20 real-life F1 racing tracks. Furthermore, CLUTR performs comparably to the non-UED attention-based CarRacing SOTA~\cite{cr-attention},  outperforming it in nine of the 20 test tracks while requiring 500X fewer environment interactions.  In navigation tasks, CLUTR outperforms PAIRED in 14 out of the 16 unseen tasks, achieving a 45\% higher solve rate. 


In summary, we make the following contributions: i) we introduce CLUTR, a novel adaptive-teacher UED algorithm derived from a hierarchical graphical model for UEDs, that augments the teacher with unsupervised task-representation learning ii) CLUTR, by decoupling task representation learning from curriculum learning, solves the long-horizon credit assignment and the combinatorial explosion problems faced by regret-based adaptive-teacher UEDs such as PAIRED. 
iii) Our experimental results show CLUTR significantly outperforms PAIRED, both in terms of generalization and sample efficiency, in two challenging domains: CarRacing and navigation.

%% file: sections/related_works.tex
\textbf{Unsupervised Curriculum Design:} \citet{paired} was the first to formalize UED and introduced the minimax regret-based UED teacher algorithm, PAIRED, with a strong theoretical robustness guarantee. However, gradient-based multi-agent RL has no convergence guarantees and often fails to converge in practice~\cite{multirl-convergence}. Pre-existing techniques like Domain Randomization (DR)~\cite{dr,dr2,dr3} and minimax adversarial curriculum learning~\cite{minimax0, minimax1} also fall under the category of UEDs. DR teacher follows a uniform random strategy, while the minimax adversarial teachers follow the maximin criteria, i.e., generate tasks that minimize the returns of the agent. POET~\cite{poet} and Enhanced POET~\cite{enhancedpoet} also approached UED, before PAIRED, using an evolutionary approach of a co-evolving population of tasks and agents. 


Recently, \citet{dcd} proposed Dual Curriculum Design (DCD): a novel class of UEDs that augments UED generation methods (e.g., DR and PAIRED) with replay capabilities. DCD involves two teachers: one that actively generates tasks with PAIRED or DR, while the other curates the curriculum to replay previously generated tasks with Prioritized Level Replay (PLR)~\cite{plr}. \citet{dcd} shows that, even with random generation (i.e., DR), updating the students only on the replayed level (but not while they are first generated, i.e., no exploratory student gradient updates as PLR) and with a regret-based scoring function, PLR can also learn minimax-regret agents at Nash Equilibrium and call this variation Robust PLR. It also introduces REPAIRED, combining PAIRED with Robust PLR. ~\citet{accel} introduces ACCEL, which improves on Robust PLR by allowing edit/mutation of the tasks with an evolutionary algorithm. Currently, random-teacher UEDs outperform adaptive-teacher UED methods.

While CLUTR and other PAIRED-variants actively adapt task generation to the performance of agents, other algorithms such as PLR generates task from a fixed-random task distribution, resulting in two categories of UED methods, i) adaptive teacher/generator based UEDs and ii) random-generator based UEDs. The existing adaptive-teacher UEDs are variants of PAIRED, which try to improve PAIRED from different aspects, but are still susceptible to the instability due to a evolving task-manifold. Unlike other PAIRED variants, CLUTR introduces a novel variational formulation with a VAE-style pretraining for task-manifold learning to solve this instability issue and can be applied, also potentially improve, any adaptive-teacher UEDs. On the other hand, random-generator UEDs focus on identifying or, prioritizing which tasks to present to the student from the randomly generated tasks, and is orthogonal to our proposed approach. 




\textbf{Representation Learning:} Variational Auto Encoders \cite{vae0,vae1,beta-vae} have widely been used for their ability to capture high-level semantic information from low-level data and generative properties in a wide variety of complex domains such as computer vision~\cite{vae-image-gen-0,vae-image,vae-image2,vae-image3}, natural language~\cite{clutr-vae, vae-sentence-1}, speech~\cite{vae-speech}, and music~\cite{vae-music}. VAE has been used in RL as well for representing image observations~\cite{vae-rl-0, vae-rl-2} and generating goals~\cite{vae-rl-1}. While CLUTR also utilizes similar VAEs, different from prior work, it combines them in a new curriculum learning algorithm to learn a latent task manifold. ~\citet{latent-goal-generation-florensa} also proposed a curriculum learning algorithm, however, for latent-space goal generation using a Generative Adversarial Network. 

%% file: sections/background.tex
\subsection{Unsupervised Environment Design (UED)}
As formalized by ~\citet{paired} UED is the problem of inducing a curriculum by designing a distribution of concrete, fully-specified environments, from an underspecified environment with free parameters. The fully specified environments are represented using a Partially Observable Markov Decision Process (POMDP) represented by $(A,O, S,\mathcal{T},\mathcal{I},\mathcal{R}, \gamma)$, where $A$, $O$, and $S$ denote the action, observation, and state spaces, respectively. $\mathcal{I} \rightarrow O$ is the observation function, $\mathcal{R}: S \rightarrow \mathbb{R}$ is the reward function, $\mathcal{T}: S \times A \rightarrow \Delta(S)$ is the transition function and $\gamma$ is the discount factor.
The underspecified environments are defined in terms of an Underspecified Partially Observable Markov Decision Process (UPOMDP) represented by the tuple $\mathcal{M} = (A,O, \Theta, S^\mathcal{M},\mathcal{T^M},\mathcal{I^M},\mathcal{R^M}, \gamma)$. $\Theta$ is a set representing the free parameters of the environment and is incorporated in the transition function as $\mathcal{T^M}: S \times A \times \Theta \rightarrow \Delta(S)$. Assigning a value to $\vec{\theta}$ results in a regular POMDP, i.e., UPOMDP +  $\vec{\theta}$ = POMDP. 
Traditionally (e.g., in ~\citet{paired} and ~\citet{dcd}) $\Theta$ is considered as a trajectory of environment parameters $\vec{\theta}$ or just $\theta$---which we call task in this paper. For example, $\theta$ can be a concrete navigation task represented by a sequence of obstacle locations. We denote a concrete environment generated with the parameter $\vec{\theta} \in \Theta$ as $\mathcal{M}_{\vec{\theta}}$ or simply $\mathcal{M}_{\theta}$. The value of a policy $\pi$ in $\mathcal{M}_{\theta}$ is defined as $V^\theta(\pi) = \mathbb{E}[\sum_{t=0}^{T} r_t \gamma^t]$, where $r_t$ is the discounted reward obtained by $\pi$ in $\mathcal{M}_{\theta}$. 
\subsection{PAIRED}
PAIRED~\cite{paired} solves UED with an adversarial game involving three players~\footnote{In the original PAIRED paper, the primary student agent was named \textit{protagonist}. Throughout this paper we refer it simply as the \textit{agent}.}: the agent $\pi_P$ and an antagonist $\pi_A$, are trained on tasks generated by the teacher $\tilde{\theta}$. PAIRED objective is:  $max_{\tilde{\theta},\pi_P}min_{\pi_A} U(\pi_P, \pi_A, \tilde{\theta}) = \mathbb{E}_{\theta\sim\tilde{\theta}}[\textsc{Regret}^\theta (\pi_P, \pi_A)] $. Regret is defined by the difference of the discounted rewards obtained by the antagonist and the agent in the generated tasks, i.e., $\textsc{Regret}^\theta (\pi_P, \pi_A) = V^\theta(\pi_A) - V^\theta(\pi_P)$. The PAIRED teacher agent is defined as $\Lambda: \Pi \rightarrow \Delta(\Theta^T)$, where $\Pi$ is a set of possible agent policies and $\Theta^T$ is the set of possible tasks. The teacher is trained with an RL algorithm with $U$ as the reward while, the protagonist and antagonist agents are trained using the usual discounted rewards from the environments. 
~\citet{paired} also introduced the flexible regret objective, an alternate regret approximation that is less susceptible to local optima. It is defined by the difference between the average score of the agent and antagonist returns and the score of the policy that achieved the highest average return.

%% file: sections/clutr_method.tex
In this section, we formally present CLUTR as a latent UED and discuss it in details. 


\subsection{Formulation of CLUTR}

\begin{wrapfigure}{r}{0.16\textwidth}
\centering
\vspace{-15pt}
\includegraphics[width=0.08\textwidth]{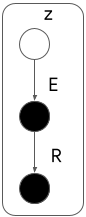}
\caption{Hierarchical Graphical Model for CLUTR}
\label{fig:graphical-model-clutr}
\vspace{-10pt}
\end{wrapfigure}

At the core of CLUTR is the latent generative model representing the latent task manifold. Let's assume that $R$ is a random variable that denotes a measure of success over the agent and antagonist agent and $z$ be a latent random variable that generates environments/tasks, denoted by the random variable $E$. We use the graphical model shown in Figure-\ref{fig:graphical-model-clutr} to formulate CLUTR. Both $E$ and $R$ are observed variables while $z$ is an unobserved latent variable. $R$ can cover a broad range of measures used in different UED methods including PAIRED and DR (Domain Randomization). In PAIRED, $R$ represents the $\textsc{Regret}$.

We use a variational formulation of UED by using the above graphical model to derive the following ELBO for CLUTR, where $VAE(z,E)$ denotes the VAE objective:

\vspace{-10pt}
\begin{align}
\label{eqn:clutr-elbo}
    ELBO &\approx VAE(z, E) - \textsc{Regret}(R, E)
\end{align}

We share the details of this derivation in Section~\ref{app:formulation} of the Appendix. The above ELBO (Eq.\ref{eqn:clutr-elbo}) defines the optimization objective for CLUTR, which can be seen as optimizing the VAE objective with a regret-based regularization term and vice versa. As previously discussed, it is difficult to train a UED teacher while jointly optimizing for both the curriculum and task representations.
Hence we propose a two-level optimization for CLUTR. First, we pretrain a VAE to learn unsupervised task representations, and then in the curriculum learning phase, we optimize for regret to generate the curriculum while keeping the VAE fixed. In Section~\ref{sec:exp-finetune-vae},  we empirically show that this two-level optimization performs better than the joint optimization of  Eq.\ref{eqn:clutr-elbo}, i.e., finetuning the VAE decoder with the regret loss during the curriculum learning phase. 

\subsection{Unsupervised Latent Task Representation Learning}
As discussed above, we use a Variational AutoEncoder (VAE) to model our generative latent task-manifold.  Aligning with ~\citet{paired} and ~\citet{dcd}, we represent task $\theta$, as a sequence of integers. For example, in a navigation task, these integers denote obstacle, agent, and goal locations. We use an LSTM-based Recurrent VAE~\cite{clutr-vae} to learn task representations from integer sequences. We learn an embedding for each integer and use cross-entropy over the sequences to measure the reconstruction error. This design choice makes CLUTR applicable to task parameterization beyond integer sequences, e.g., to sentences or images. To train our VAEs, we generate random tasks by uniformly sampling from $\Theta^T$, the set of possible tasks. Thus, we do not require any interaction with the environment to learn the task manifold. Such unsupervised training of the task manifold is practically very useful as interactions with the environment/simulator are much more costly than sampling. Furthermore, we sort the input sequences, fully or partially, when they are permutation invariant, i.e., essentially represent a set. By sorting the training sequences, we avoid the combinatorial explosion faced by other adaptive UED teachers. 

\subsection{CLUTR}
\begin{algorithm}[!h]
\begin{algorithmic}[1]
\STATE Pretrain VAE with randomly sampled tasks from $\Theta$
\STATE Randomly initialize Agent $\pi^{P}$, Antagonist $\pi^{A}$, and Teacher $\tilde{\Lambda}$;
\REPEAT
    \STATE Generate latent task vector: $z \sim \mathcal{Z}$ from the teacher 
    \STATE Create POMDP $M_{\theta}$ where $\theta=\mathcal{G}(z)$ and $\mathcal{G}$ is the VAE decoder function
    \STATE Collect Agent trajectory $\tau^{P}$ in $M_{\theta}$. Compute: $U^{\theta}(\pi^{P})$ = $\sum_{i=0}^{T} r_{t} \gamma^{t}$
    \STATE Collect Antagonist trajectory $\tau^{A}$ in $M_{\theta}$. Compute: $U^{\theta}(\pi^{A})$ = $\sum_{i=0}^{T} r_{t} \gamma^{t}$
    \STATE Compute: $\textsc{Regret}^{\theta} (\pi^{P},\pi^{A}) = U^{\theta} (\pi^{A}) - U^{\theta} (\pi^{P})$
    \STATE Train Protagonist policy $\pi^{P}$ with RL update and reward $R(\tau^{P}) = U^{\theta}(\pi^{P})$
    \STATE Train Antagonist policy $\pi^{A}$ with RL update and reward $R(\tau^{A}) = U^{\theta}(\pi^{A})$
    \STATE Train Teacher policy $\tilde{\Lambda}$ with RL update and reward $R(\tau^{\tilde{\Lambda}}) = \textsc{Regret}$
\UNTIL{\textit{not converged}}
\end{algorithmic}
\caption{CLUTR}
\label{algo:CLUTR}
\end{algorithm}

We define CLUTR following the objective given in Eq.~\ref{eqn:clutr-elbo}. CLUTR uses the same curriculum objective as PAIRED, $\textsc{Regret}(R, E) = \textsc{Regret}^\theta (\pi_P, \pi_A)$ where, $\theta$ denotes a task, i.e., a concrete assignment to the free parameters of the environment $E$. Unlike PAIRED teacher, which generates $\theta$ directly, the CLUTR teacher policy is defined as $\Lambda: \Pi \rightarrow \Delta(\mathcal{Z})$, where $\Pi$ is a set of possible agent policies and $\mathcal{Z}$ is as the latent space. Thus, the CLUTR teacher is a latent environment designer, which samples random $z$ and $\theta$ is generated by the VAE decoder function $\mathcal{G}: \mathcal{Z} \rightarrow \Theta$. We present the outline of the CLUTR in Algorithm~\ref{algo:CLUTR}. CLUTR outline is very similar to PAIRED, differing only in the first two lines of the main loop to incorporate the latent space.  

Now we discuss a couple of additional properties of CLUTR compared to other adaptive-teacher UEDs, i.e., PAIRED and REPAIRED. First, CLUTR teacher samples from the latent space $\mathcal{Z}$ and thus generates a task in a single timestep. Note that this is not possible for other adaptive UED teachers, as they operate on parameter space and generate one task parameter at a time, conditioned on the state of the partially-generated task so far. Furthermore, Adaptive-teacher UEDs typically observe the state of their partially generated task to generate the next parameters. Hence they require designing different teacher architectures for environments with different state space. CLUTR teacher architecture, however, is agnostic of the problem domain and does not depend on their state space.   Hence the same architecture can be used across different environments.

\subsection{CLUTR in the context of contemporary UED method landscape}\label{sec:contemporary-ueds}
\begin{table*}[]
\centering
\begin{tabular}{|l|c|c|c|c|}
\hline
Algorithm  & \begin{tabular}[c]{@{}c@{}}Task \\ Representation  Learning\end{tabular} & \begin{tabular}[c]{@{}c@{}}Teacher \\ Model\end{tabular} & \begin{tabular}[c]{@{}c@{}}UED \\ Method\end{tabular} & \begin{tabular}[c]{@{}c@{}}Replay\\  Method\end{tabular} \\ \hline
DR         & \multirow{4}{*}{-}        & \multirow{4}{*}{Random}              & \multirow{3}{*}{DR}                                   & -                                                        \\  \cline{5-5} 
PLR        &                           &                                      &                                                       & PLR                                                      \\ \cline{5-5} 
Robust PLR &                           &                                      &                                                       & \multirow{2}{*}{Robust PLR}                              \\ \cline{4-4}
ACCEL      &                                                                   &                                      & DR + Evolution                                        &                                                          \\ \hline
PAIRED     & \multirow{2}{*}{Implicit via RL}                                    & \multirow{3}{*}{Learned}             & \multirow{3}{*}{Regret}                               & -                                                        \\  \cline{5-5} 
REPAIRED   &                                                                   &                                      &                                                       & Robust PLR                                               \\ \cline{1-2} \cline{5-5} 
CLUTR      & \begin{tabular}[c]{@{}c@{}}Explicit via  \\ Unsupervised Generative Model\end{tabular}                                        &                                      &                                                       & -                                                        \\ \hline
\end{tabular}
\caption{A comparative characterization of contemporary UED methods}
\label{tab:ued_classification}
\end{table*}
As discussed in Section~\ref{sec:related}, contemporary UED methods can be characterized by their i) teacher type (random/fixed or, learned/adaptive) and, ii) the use of replay. To clearly place CLUTR in the context of contemporary UEDs, we discuss another important aspect of curriculum learning algorithms: how the task manifold is learned. The random-generator UEDs (e.g., DR, PLR) do not learn a task manifold. Regret-based adaptive-teachers, i.e., PAIRED and REPAIRED, learn an implicit (e.g., the hidden state of the teacher LSTM) task-manifold---from scratch---but it is not utilized explicitly. It is trained via RL, based on the regret estimates of the tasks they generate. Hence, these task-manifolds depend on the quality of the estimates, which in turn depends on the overall health of the multi-agent RL training. Furthermore, they do not take into account the actual task structures. In contrast, CLUTR introduces an explicit task-manifold modeled with VAE, that can represent a local neighborhood structure capturing the similarity of the tasks, subject to the parameter space being used. Hence, similar tasks (in terms of parameterization) would be placed nearby in the latent space. Intuitively this local neighborhood structure should facilitate the teacher to navigate the manifold effectively. The above discussion illustrates that CLUTR along with PAIRED and REPAIRED form a category of UEDs that generates tasks based on a learned task-manifold, orthogonal to the random generation-based methods, while CLUTR being the only one utilizing an unsupervised generative task manifold. Table~\ref{tab:ued_classification} summarizes the similarity and differences.

%% file: sections/experiments.tex
In this section, we evaluate CLUTR in two challenging domains: i) Pixel-Based Car Racing with continuous control and dense rewards, and ii) partially observable navigation tasks with discrete control and sparse rewards. We compare CLUTR primarily with PAIRED to analyze its impact on improving adaptive-teacher UED algorithms, experimenting with two commonly used regret objectives: standard and flexible.  As discussed in Section~\ref{sec:related} and ~\ref{sec:contemporary-ueds}, there are other random-generator and adaptive-teacher UEDs employing techniques complimentary or orthogonal to our approach. For completeness, we compare CLUTR with such existing UED methods in Section~\ref{app:cr-detailed-comparison}~and ~\ref{app:mg-details} in the Appendix.



We then empirically investigate the following hypotheses: \\
\textbf{H1}: Simultaneous learning of latent task manifold and curriculum  degrades performance (Section~\ref{sec:exp-finetune-vae}) \\
\textbf{H2}: Training VAE on sorted data solves the combinatorial explosion problem. (Section~\ref{sec:exp-sorting-vae-data}) 

At last, we analyze CLUTR curriculum in multiple aspects while comparing it with PAIRED to have a closer understanding. Full details of the environments, network architectures, training hyperparameters, VAE training and further details are discussed in the Appendix.

\subsection{CLUTR Performance on Pixel-Based Continuous Control CarRacing Environment}
\label{sec:clutr-carracing-main}
The CarRacing environment~\cite{dcd,openai_gym} requires the agent to drive a full lap around a closed-loop racing track modeled with B\'{e}zier Curves~\cite{bezier} of up to 12 control points.  Both CLUTR and PAIRED were trained for 2M timesteps for flexible regret objective and for 5M timesteps for the standard regret objective experiments. We train the VAE on 1 million randomly generated tracks for 1 million gradient updates. Note that only one VAE was trained and used for all the experiments (10 independent runs, both objectives). We evaluate the agents on the F1 benchmark~\cite{dcd} containing 20 test tracks modeled on real-life F1 racing tracks. These tracks are significantly out of distribution than any tracks that the UED teachers can generate with just 12 control points. Further details on the environment, network architectures, VAE training, and detailed experimental results with analysis can be found in Section~\ref{app:environment},~\ref{app:network-architecture},~\ref{app:vae-training},~\ref{app:carracing-exp-detailed} of the Appendix, respectively. 

\begin{figure}[!htbp]
     \centering
     \includegraphics[width=0.3\textwidth]{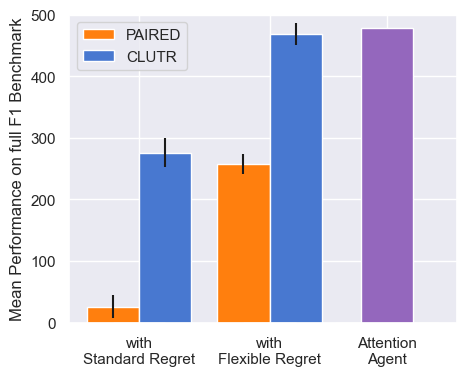}
     \caption{Comparison on the F1 Benchmark comprising 20 tracks modeled on real-life F1 racing tracks collected from 10 independent runs. CLUTR achieves 10.6X and 82\%  higher returns than PAIRED with standard and flexible regret objectives, respectively. CLUTR also performs comparably to the attention-based non-UED CarRacing SOTA, while requiring 500X fewer environment interactions.}
    \label{fig:cr-final-comp-main}
\end{figure}

\begin{figure}[htbp!]
     \centering
      \begin{subfigure}[h]{0.23\textwidth}
         \centering
         \includegraphics[width=\textwidth]{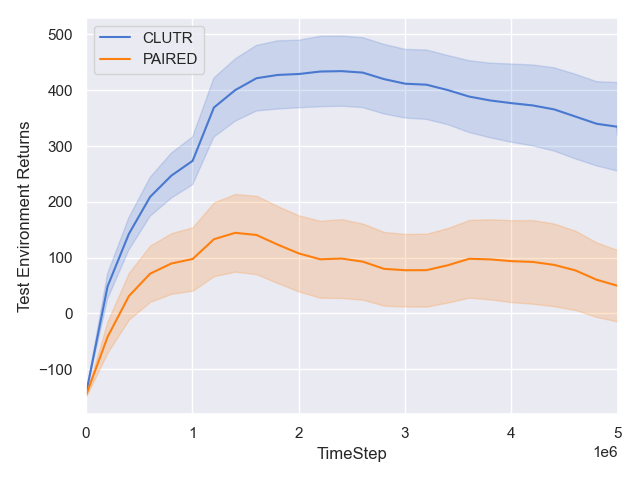}
         \caption{with Standard Objective}
         \label{fig:cr-eval-during-training-std}
     \end{subfigure}   
     \hfill
    \begin{subfigure}[h]{0.23\textwidth}
         \centering
         \includegraphics[width=\textwidth]{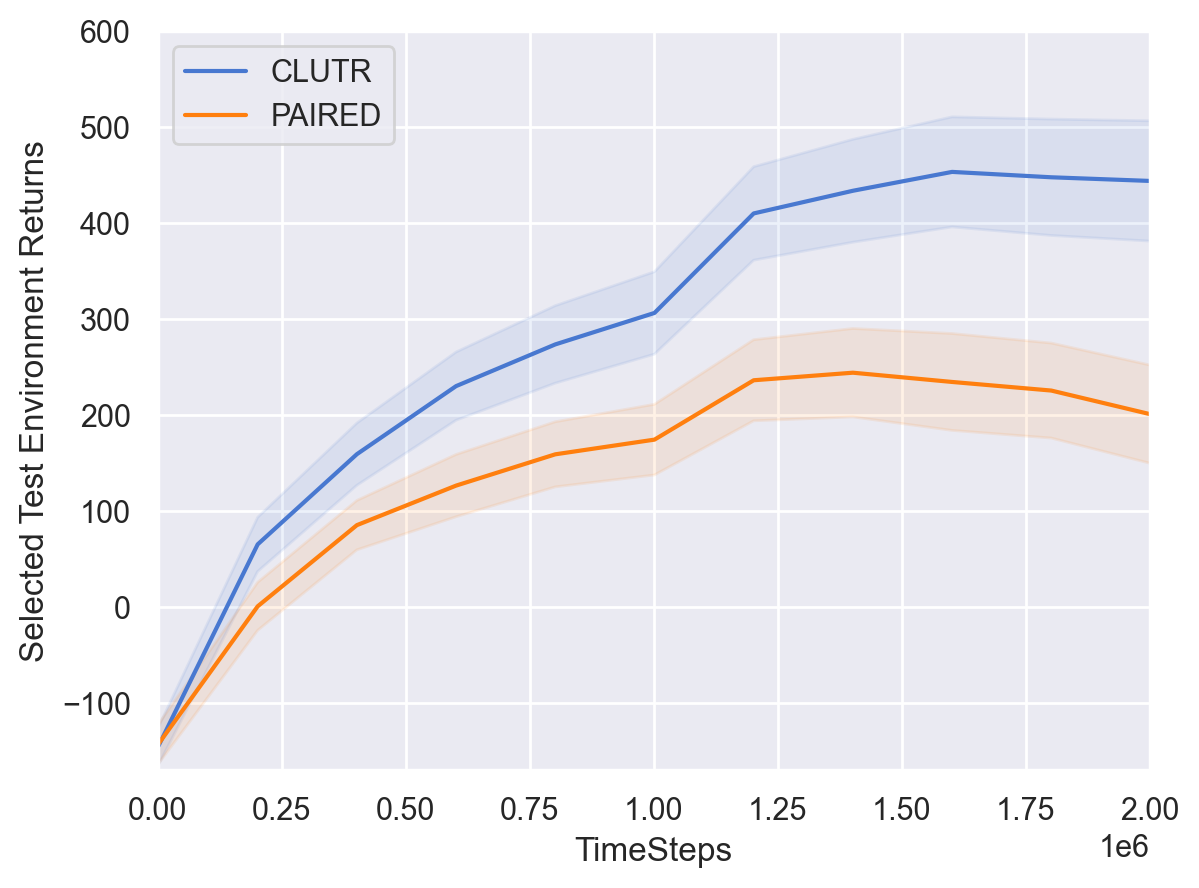}
         \caption{with Flexible Objective}
         \label{fig:cr-eval-during-training-flx}
     \end{subfigure}
    \caption{Zero-shot generalization over the course of training by periodic evaluation on a subset of three F1 tracks: Singapore, Germany, and Italy. CLUTR indicate significantly better sample efficiency than PAIRED.}
    \label{fig:cr-eval-during-train}
\end{figure}

Figure~\ref{fig:cr-final-comp-main} shows the mean return obtained by CLUTR and PAIRED on the full F1 benchmark, on. We independently experimented with both the standard and flexible regret objectives. We notice that PAIRED performs miserably with standard regret in these tasks. However, implementing CLUTR or changing to the flexible regret objective, improves the performance considerably. Furthermore, CLUTR with flexible regret results in much better performance, comparable to the non-UED attention-based SOTA for CarRacing~\cite{cr-attention}, despite not using a self-attention policy and training on 500X fewer environment interactions, while outperforming it on nine of the 20 F1 tracks (See Table~\ref{tab:clutr-cr-full} in Appendix). We also note, CLUTR improves PAIRED irrespective of the choice of the regret objectives: achieving 10.6X and 82\% higher returns with standard and flexible regret objectives, respectively and outperforming PAIRED on each of the 20 F1 tracks (See Table~\ref{tab:clutr-cr-full}). Figure~\ref{fig:cr-eval-during-train} illustrate the agents' generalization capabilities during training, by periodically evaluating them on a subset of three unseen F1 tracks: Singapore, Germany, and Italy, which are selected aligning with~\citet{dcd}. Based on these environments, CLUTR shows significantly better trends of sample efficiency, achieving better generalization with significantly fewer environment interactions compared to PAIRED. 
Furthermore, CLUTR (with flexible regret) emerges as the best adaptive-teacher UED for CarRacing outperforming the other adaptive-teacher UED: REPAIRED and random-generator UEDs: DR, and PLR by 58\%, 38\% and 16\%, repectively. CLUTR is also the only adaptive-teacher UED that outperforms the random-teacher UED methods. CLUTR falls short (by 14\%) only to Robust PLR---a random generator dual-curriculum UED with replay and stop-gradient capabilities---a method fundamentally different than ours or, PAIRED. Further discussion can be found in Section~\ref{app:cr-detailed-comparison}. 

\subsection{CLUTR Performance on Partially Observable Navigation Tasks on MiniGrid}
\label{sec:clutr-mg-main}

We now compare CLUTR with PAIRED on the popular MiniGrid environment, originally introduced by~\cite{minigrid} and adopted by~\cite{paired} for UEDs, for both standard and flexible regret objectives. In these navigation tasks, an agent explores a grid world to find the goal while avoiding obstacles and receives a sparse reward upon reaching the goal. For flexible regret experiment, we generated 10 million random grids to train the VAE, with the obstacle locations sorted, and the number of obstacles  uniformly varying from zero to 50, aligning with~\cite{paired}. The standard regret experiment uses a similar but smaller dataset of 1 million grids. Note that the results reported in the original PAIRED paper are obtained after 3 billion timesteps of training, while we train PAIRED and CLUTR for 250M and 500M timesteps (5 independent runs), for flexible and standard regret objectives, respectively. We evaluate on a testset of 16 novel navigation tasks from~\citet{paired}.



\begin{figure}[!htbp]
     \centering
     \includegraphics[width=0.35\textwidth]{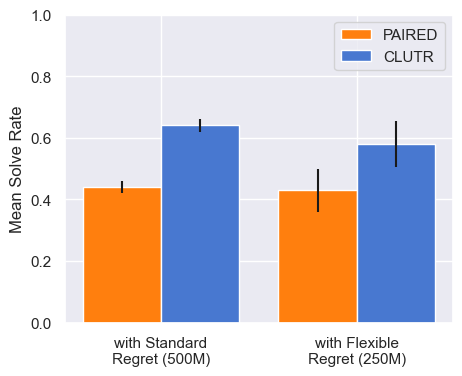}
     \caption{Mean solve rate on the test dataset comprising 16 novel nagivation tasks from 5 independent runs. CLUTR achieves 45\% and 35\%  higher solve rate than PAIRED, with standard and flexible regret objectives, respectively.}
    \label{fig:mg-final-comp-main}
\end{figure}

\begin{figure}[htbp!]
     \centering
      \begin{subfigure}[h]{0.24\textwidth}
         \centering
         \includegraphics[width=\textwidth]{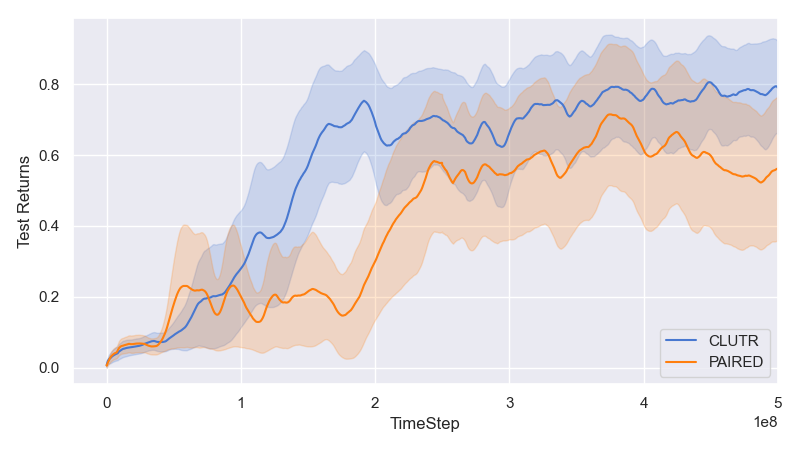}
         \caption{with Standard Objective}
         \label{fig:mg-eval-during-training-std}
     \end{subfigure}   
     \hfill
    \begin{subfigure}[h]{0.2\textwidth}
         \centering
         \includegraphics[width=\textwidth]{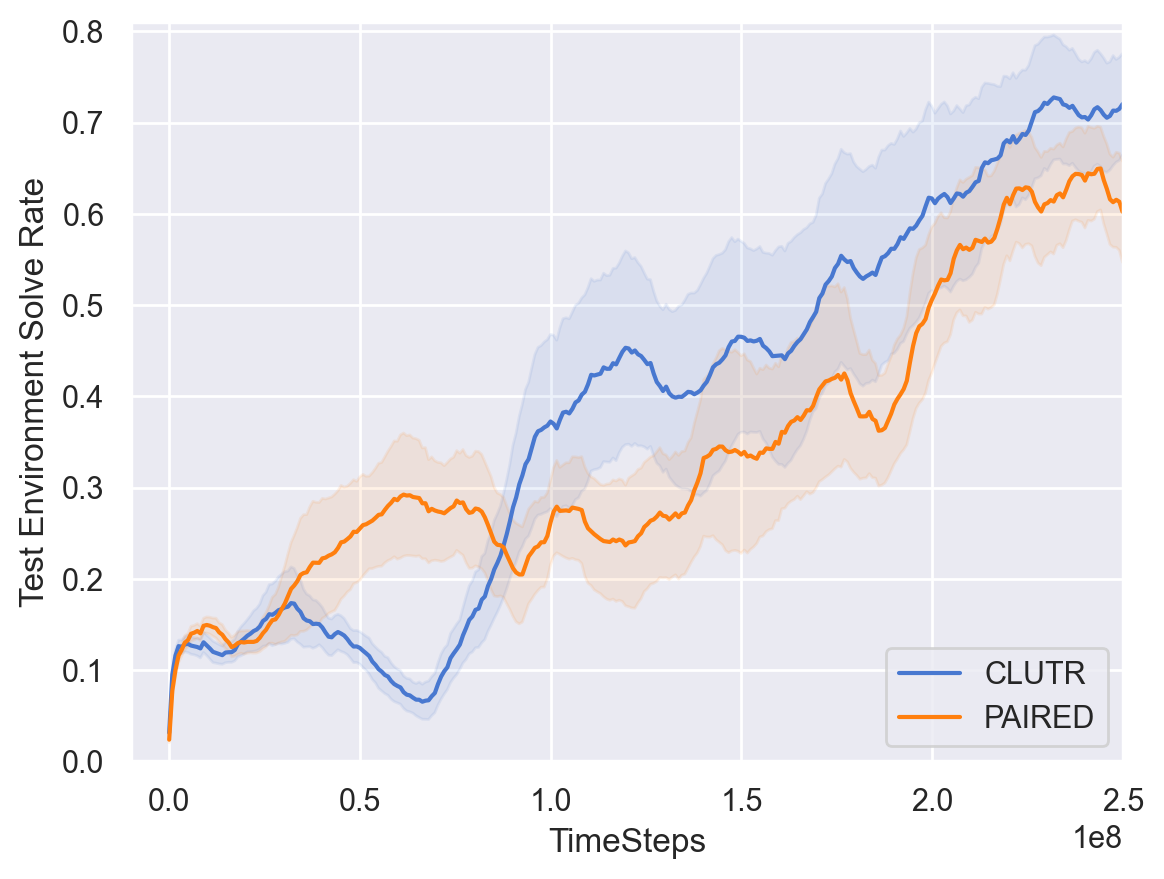}
         \caption{with Flexible Objective}
         \label{fig:mg-eval-during-training-flx}
     \end{subfigure}
    \caption{Agent solved rate on the 16 unseen grids from~\citet{paired} during training. CLUTR shows better sample efficiency and generalization than PAIRED. The results show an average of 5 independent runs..}
    \label{fig:mg-eval-during-train}
\end{figure}

Figure~\ref{fig:mg-final-comp-main} shows the mean solve rate obtained by CLUTR and PAIRED on the test dataset. CLUTR improves PAIRED irrespective of the choice of the regret objectives: 45\% and 35\%  higher solve rate than PAIRED outperforming on 14 and 13 individual test grids out of 16 (See Figure~\ref{fig:all-mg-pairedp-vs-clutrp} and Figure~\ref{fig:all-mg-pairedfp-vs-clutrfp} in Section~\ref{app:mg-details} for details), with standard and flexible regret objectives, respectively. Figure~\ref{fig:mg-eval-during-train} plot solve rate on all the 16  test grids during training for flexible objective and a subset of four grids, namely, Sixteen Rooms, Sixteen Rooms with Fewer Doors, Labyrinth, and Large Corridor, for standard objective. We see CLUTR, though showing an initial dip for flexible objective, shows better sample efficiency by achieving a higher solve rate earlier than PAIRED.

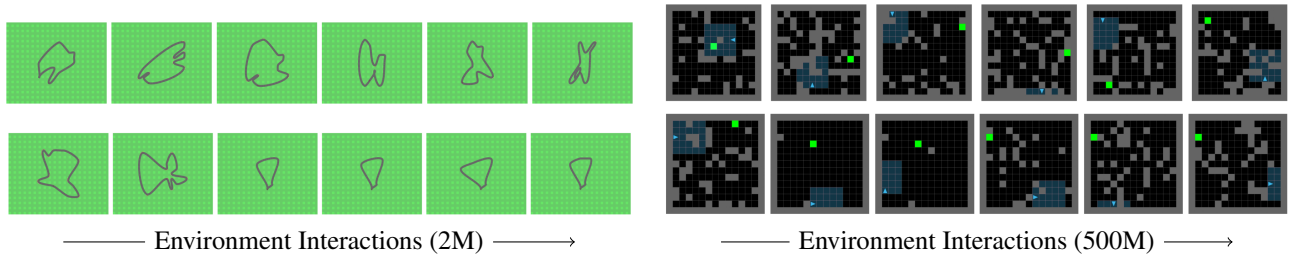
\begin{figure*}[!tb]
     \centering
     \begin{subfigure}[b]{0.49\textwidth}
         \centering
         \includegraphics[width=\textwidth]{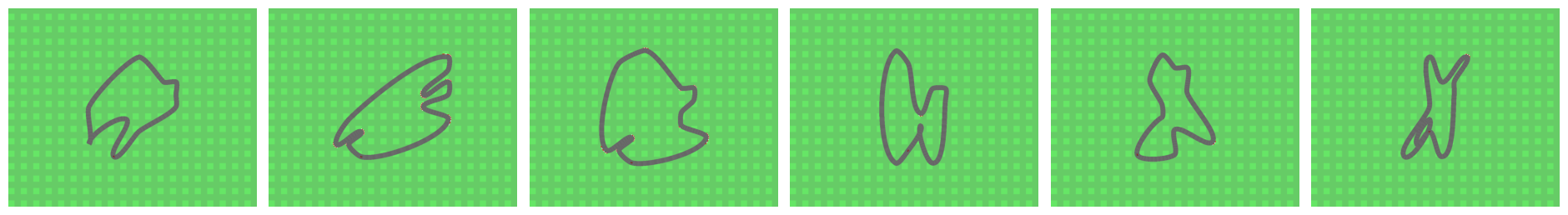}
     \end{subfigure}
     \hfill
     \begin{subfigure}[b]{0.49\textwidth}
         \centering
         \includegraphics[width=\textwidth]{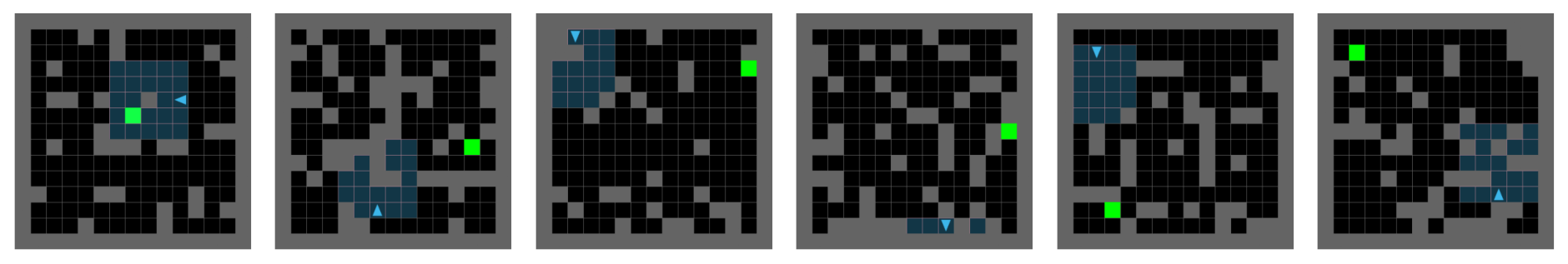}
     \end{subfigure}
     \centering
    \begin{subfigure}[b]{0.49\textwidth}
         \centering
         \includegraphics[width=\textwidth]{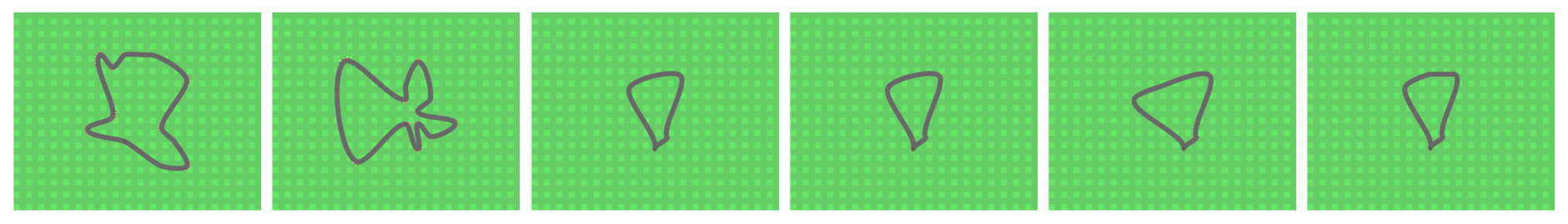}
     \end{subfigure} 
     \hfill
      \begin{subfigure}[b]{0.49\textwidth}
         \centering
         \includegraphics[width=\textwidth]{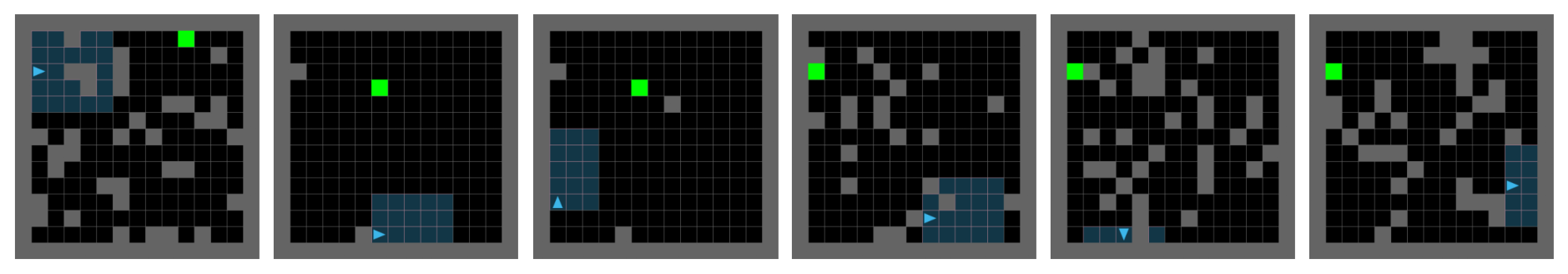}
     \end{subfigure} 
     \centering
      \begin{subfigure}[b]{0.49\textwidth}
         \centering
            \begin{tikzpicture}
            \coordinate (A) at (-3.5,0);
            \coordinate (B) at ( 3.3,0);
            \path (A) -- node (success) {Environment Interactions (2M)} (B);
            \draw[->] (A) -- (success) -- (B);
          \end{tikzpicture}
     \end{subfigure}
    \hfill
      \begin{subfigure}[b]{0.49\textwidth}
         \centering
            \begin{tikzpicture}
\coordinate (A) at (-3.5,0);
            \coordinate (B) at ( 3.3,0);
            \path (A) -- node (success) {Environment Interactions (500M)} (B);
            \draw[->] (A) -- (success) -- (B);
          \end{tikzpicture}
     \end{subfigure}
     
    \caption{Example tracks(left) and grids(right) generated by CLUTR(top) and PAIRED(bottom) uniformly sampled at different stages of training. The training progresses from left to right. PAIRED seems to generate over simplified tasks for substantial amount of time hampering agent learning. CLUTR generates interesting tasks throughout.}
    \label{fig:curriculum}
\end{figure*}

\subsection{Learning task manifold and curriculum: Joint vs Two-staged Optimization} \label{sec:exp-finetune-vae} 

\begin{figure}[!htbp]
     \centering
     \vspace{-10pt}
     \includegraphics[width=0.4\textwidth]{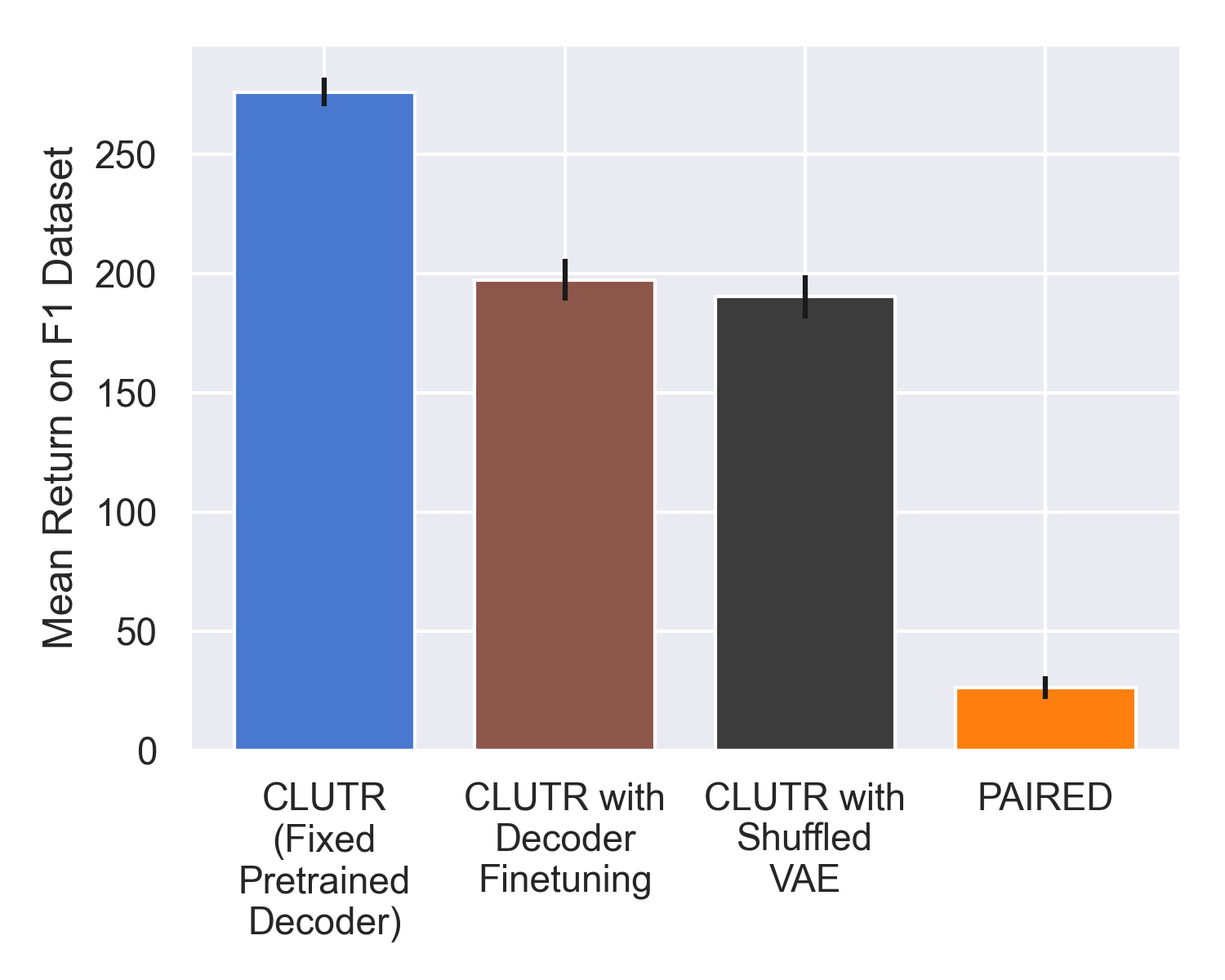}
     \caption{Impact of i) joint vs two-staged optimization of the task manifold and ii) using a `Shuffled' VAE, trained on a larger shuffled dataset. The leftmost column shows the default CLUTR performance---i.e., using a pretrained decoder (VAE) trained on sorted training data, kept fixed during the curriculum learning phase---with standard regret objective for CarRacing. Allowing the decoder to finetune with the regret loss results in a 29\% performance drop and the use of Shuffled VAE shows a drop of 31\%. These performance drops empricially justify our hypotheses \textbf{H1} and \textbf{H2}. Also, CLUTR with decoder finetuning and Shuffled VAE still outperform PAIRED, with 7.6X and 7.3X better returns, respectively.}
    \label{fig:mean-test-clutr-vs-clutrsh-vs-clutrft}
\end{figure} 

We hypothesized that learning the task representation and the curriculum simultaneously results in a difficult learning problem due to the non-stationarity of the process. To test this, we conduct an experiment in which we allow finetuning our pretained decoder with the regret loss during the curriculum learning phase. This experiment, namely `CLUTR with Decoder Finetuning', shows a 29\% performance drop in the CarRacing domain with the standard regret objective (Figure~\ref{fig:mean-test-clutr-vs-clutrsh-vs-clutrft}). Similarly, we see a drop of 10\% in case of flexible regret further justifying our hypothesis (See Section~\ref{app:finetune-vae-cr-flexible} for details). As a side note, the smaller drop in the later case indicates that flexible objective mitigates some of the instability problem too. Finally, even with decoder finetuning, CLUTR achieves 7.6X and 65\% improvement over PAIRED, for standard and flexible regret respectively--- indicating the benefits of pretrained decoupled latent task space. 
The above experimental results thus empirically validates our hypothesis that keeping the pretrained task manifold fixed during curriculum learning helps solving the instability problem. 

\subsection{Impact of sorting VAE data on solving Combinatorial Explosion}\label{sec:exp-sorting-vae-data} 

We hypothesized that training a VAE on sorted sequences can solve the combinatorial explosion problem. To test this, we conduct an experiment, `CLUTR with Shuffled VAE', in which we train CLUTR with an alternate VAE---trained 5X longer on a non-sorted and 10X bigger version of the original dataset. This experiment shows a 31\% performance drop in the CarRacing domain as seen in Figure~\ref{fig:mean-test-clutr-vs-clutrsh-vs-clutrft}, empirically validating our hypothesis. On another note, CLUTR with Shuffled VAE still shows a 7.3X improvement over PAIRED. This indicates that, even when the task manifold is `suboptimal', a fixed and pretrained task-manifold, i.e., the decoupling of task representation and curriculum learning, helps solving the learning instability and combinatorial explosion problem faced by PAIRED. Further details of this experiment are discussed in Section ~\ref{app:shuffled-vae} of the Appendix. 

\begin{figure}[htbp!]
     \centering
     \begin{subfigure}[h]{0.23\textwidth}
         \centering
         \includegraphics[width=\textwidth]{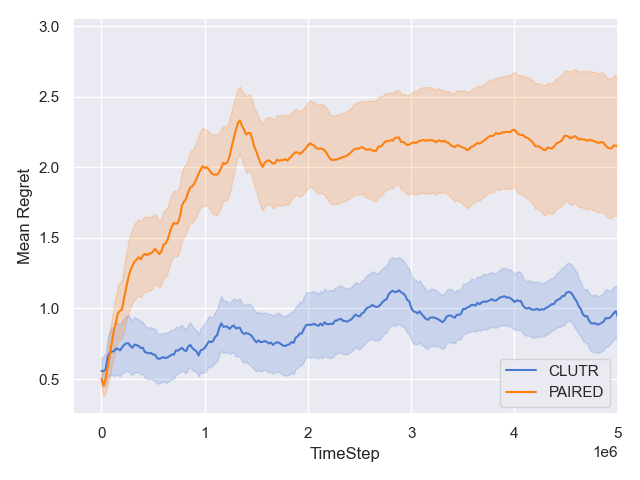}
         \caption{Mean Regret - CarRacing}
         \label{fig:regret-cr-fp}
     \end{subfigure}
      \begin{subfigure}[h]{0.24\textwidth}
         \centering
         \includegraphics[width=\textwidth]{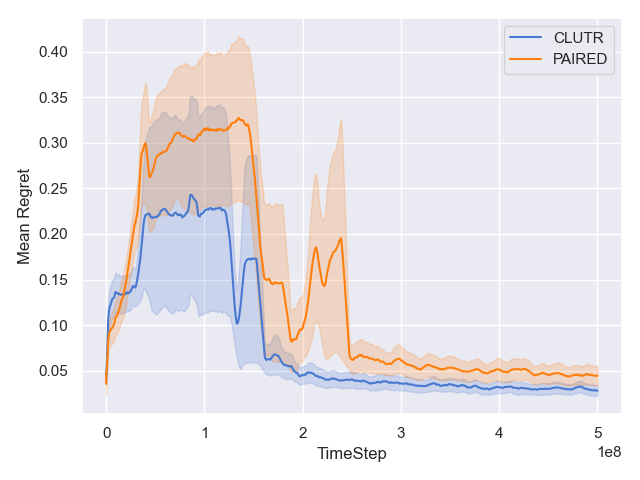}
         \caption{Mean Regret - Navigation}
         \label{fig:regret-mg50-fp}
     \end{subfigure}    
    \caption{Mean standard regret during training. CLUTR shows a smaller regret value indicating a smaller performance gap between the agent and the antagonist, compared to PAIRED. }
    \label{fig:regret-main-fp}
\end{figure}

\subsection{Analysis of the Curriculum: CLUTR vs PAIRED}
\label{sec:exp-teacher-training}
In Section~\ref{sec:clutr-carracing-main} and \ref{sec:clutr-mg-main} we discussed how CLUTR outperforms PAIRED, both in terms of sample efficiency and generalization, suggesting CLUTR induces a significantly more effective curriculum than PAIRED.  For better understanding of CLUTR curriculum, in Figure~\ref{fig:regret-main-fp} we analyze the mean regret---the performance gap between the agent and the adversary---on the teacher-generated curricula for both CarRacing and navigation tasks. 

CLUTR and PAIRED show similar regret patterns, which is not surprising as both optimize regret using the same criteria. However, CLUTR converges to a smaller regret value; faster than PAIRED. From a curriculum learning perspective, smoother training is expected with tasks that are `slightly' harder than the agent can already solve or, can obtain `slightly' better returns. In practice, both the agent and the antagonist are trained in the same training data and context e.g., the same hyper-parameters, architecture, differing only by their random initial weights. Hence, a lower regret implies that the teacher is generating tasks at the frontier of the agents' capabilities, which are either slightly harder than the agent should be able to solve (because antagonist is solving them) or, the tasks in which antagonist is performing slightly better. On the other hand, higher regret values can result from generating tasks which are biased towards the strength or, idiosyncracy of only one of the agents, which might not be useful for generalization. In fact, PAIRED has shown to over exploit the relative strength of the antagonist for CarRacing (~\citet{dcd}), inducing curriculum showing high regret but poor generalization. Furthermore, a high regret can also imply the antagonist becoming significantly better than agent, which may lead to the teacher not having enough incentive to generate novel and diverse tasks, harming agent learning. Hence the lower regret value, might indicate that CLUTR is identifying the frontier of agents' capabilities better than PAIRED and thus inducing a more effective curriculum for training the student agents, as supported by the empirical performance. 

Figure~\ref{fig:curriculum} shows snapshots of CLUTR and PAIRED generated curriculums as training progress. We notice, PAIRED generates over-simplified tasks for substantial amount of time, which might hamper its generalization and sample efficiency. On the other hand, CLUTR doesnt seem to start with overly-simplistic tasks, rather generates tasks with a wide range of difficulty throughout. Section~\ref{app:curriculum-analysis-mg} shares detailed analysis supporting the above observation and further insights.

%% file: sections/conclusions.tex
In this work, we introduce CLUTR, an unsupervised latent space adaptive-teacher UED method that augments adaptive UED teachers with a pretrained latent task manifold to decouple task representation learning from curriculum learning. CLUTR first trains a recurrent VAE from random tasks to learn the latent task manifold and then employs a regret-based adaptive-teacher to induce the curriculum. Through this decoupling, CLUTR solves the long-horizon credit assignment and the combinatorial explosion problems faced by regret-based adaptive-teacher UED methods. Our experimental results show strong empirical evidence supporting the effectiveness of our proposed approach. 




Even though CLUTR and other regret-based UEDs empirically show good generalization on human-curated complex transfer tasks, they rarely can generate human-level task structures during training. An interesting direction would be to enable UED algorithms to generate realistic tasks. Another important direction would be to reduce the gap between the theoretical and practical aspects of regret-based multi-agent UED algorithms, which are subject to the quality of regret estimates and multi-agent RL training. At last, random generator algorithms like Robust PLR or even, DR have been shown to perform better than adaptive-teacher approaches like CLUTR or PAIRED. An interesting direction would be to investigate the conditions/environments under which a random generator performs better than an adaptive generator and vice versa. At last, we are excited about latent-space curriculum design and  hope our work will encourage further research in this domain.


%% file: sections/appendices.tex
\section{Additional Details of CLUTR}

\subsection{CLUTR Objective Derivation}
\label{app:formulation}
\begin{wrapfigure}{r}{0.22\textwidth}
\centering
\includegraphics[width=0.08\textwidth]{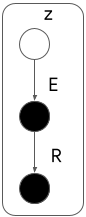}
\caption{Hierarchical Graphical Model for CLUTR}
\label{fig:graphical-model-clutr2}
\end{wrapfigure}

We use a hierarchical graphical model to formulate the latent environment design problem.
Let's assume that $R$ is a random variable that denotes a measure of success defined using the protagonist and antagonist agents and $z$ be a latent random variable.
We use the graphical model in Figure-\ref{fig:graphical-model-clutr2} where $z$ generates an environment $E$ and $R$ is the success defined over $E$.
Both $E$ and $R$ are observed variables while $z$ is an unobserved variable.
$R$ covers a broad range of measures used in different UED methods including PAIRED and DR (Domain Randomization). In PAIRED, $R$ represents the $\textsc{Regret}$ as the difference of returns between the antagonist and protagonist agents and it depends on the environments that the agents are evaluated on.

We use a variational formulation of UED by using the above graphical model.
We first define the variational objective as the KL-divergence between an approximate posterior distribution and true posterior distribution over latent variable $z$, 
\begin{align*}
    D_{KL}(q(z) | p(z | R, E)) &= E_{z \sim q(z)}[logq(z)] - E_{z \sim q(z)}[log p(z | R, E)] \\
    &= E_{z \sim q(z)}[log q(z)] - E_{z \sim q(z)}[log p(R, E, z)] + logp(R, E) \\
\end{align*}
where both $R$ and $E$ are given.

Next, we write the ELBO,
\begin{align*}
    ELBO &= E_{z \sim q(z)}[log q(z)] - E_{z \sim q(z)}[log p(R, E, z)] \\
    &= E_{z \sim q(z)}[log q(z)] - E_{z \sim q(z)}[log p(R | E)p(E | z)p(z)] \\
    &= E_{z \sim q(z)}[log q(z)] - E_{z \sim q(z)}[log p(z)] - E_{z \sim q(z)}[log p(E | z)] - E_{z \sim q(z)}[log p(R | E)] \\
    &= E_{z \sim q(z)}[log \frac{q(z)}{p(z)}] - E_{z \sim q(z)}[log p(E | z)] - log p(R | E) \\
    &= D_{KL}(q(z) | p(z)) - E_{z \sim q(z)}[log p(E | z)] - log p(R | E) \\
    &= VAE(z, E) - log p(R | E) \\
\end{align*}

We can also induce an objective that includes minimax $\textsc{Regret}$.
Let $R$ be distributed according to an exponential distribution, 
$p(R|E) \propto exp($\textsc{Regret}$(\pi_P, \pi_A | E))$, 

we derive,
\begin{align*}
    ELBO &\approx VAE(z, E) - \textsc{Regret}(R, E) \\
\end{align*}
where the normalizing factor is ignored. 

\subsection{Robustness Guarantees}
CLUTR essentially proposes including a pretrained latent space within the teacher/generator. From the teacher's perspective, the difference is while the PAIRED teacher starts from randomly initialized weights, CLUTR starts from the pretrained weights. Thus, CLUTR does not impose new assumptions on possible teacher policies. Furthermore, CLUTR does not change any other specifics of the underlying PAIRED algorithm. Hence, CLUTR holds the same theoretical robustness guarantees provided by PAIRED.  

In practice, both CLUTR and PAIRED deviate from these theoretical guarantees. For example, both algorithms approximate the regret value, which is the case for other regret-based UEDs such as Robust PLR and REPAIRED (\cite{dcd}). Also, the robustness guarantee depends on reaching the Nash equilibrium of the multiagent adversarial game. However, gradient-based multi-agent RL has no convergence guarantees and often fails to converge in practice(\cite{multirl-convergence}). We also note that, by introducing the latent space, CLUTR VAE might not have access to the full task space due to practical limitations on training, e.g., the training dataset not having all possible tasks. However, when the decoder is allowed to be finetuned, CLUTR will have access to the full task space, similar to PAIRED. Our empirical results (discussed in Section~\ref{sec:exp-finetune-vae}) suggest that keeping the pretrained decoder fixed performs better than finetuning it, so we kept it fixed for our main experiments. We also want to mention, when the flexible objective is used, CLUTR (and PAIRED) does not hold the robustness guarantee as it changes the dynamics of the underlying game between the teacher and the agents, even though flexible regret works better in practice. 





\section{Training Details}

\subsection{Environment Details}
\label{app:environment}
\begin{figure}
    \centering
    \includegraphics[width=\textwidth]{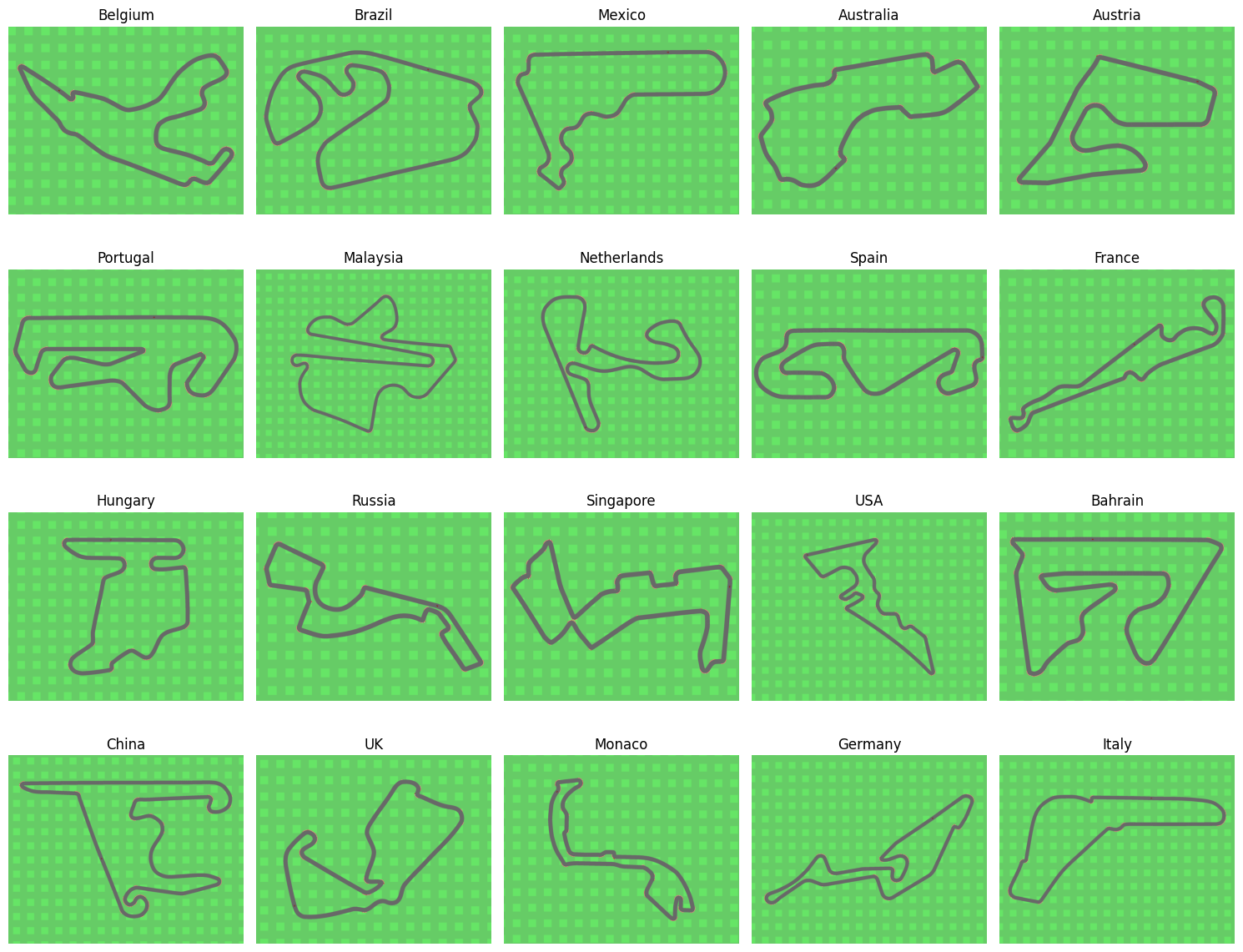}
    \caption{Snapshots of the test tracks in F1 benchmark}
    \label{app:fig-cr-test-images}
\end{figure}

\textbf{Car Racing}: The CarRacing environment was originally proposed by OpenAI Gym~\cite{openai_gym}, and later has been reparameterized by~\cite{dcd} with B\'{e}zier Curves(~\cite{bezier}) for UED algorithms. This environment requires the agents to drive a full lap around a closed-loop track. The track is defined by a  B\'{e}zier Curve modeled with a sequence of upto 12 arbitrary control points, each spaced within a fixed radius $B/2$ of the center of the $B\times B$ field. This sequence of control points can uniquely identify a track, subject to a set of predefined curvature constraints~\cite{dcd}. The control points are encoded in a $10\times 10$ grid---a discrete downsampled version of the racing track field. Each control point hence is a integer denoting a cell of the grid and the cell coordinates are upscaled to match the original scale of the field afterwards. This ensures no two control points are too close together, preventing areas of excessive track overlapping. The track consists of a sequence of $L$ polygons and the agent receives a reward of $1000/L$ upon visiting each unvisited polygon and a penalty of $-0.1$ at each time step to incentivize completing the tracks faster. Episodes terminate if the agent drives too far off-track but is not given any additional penalty. The agent controls a 3 dimensional continuous action space corresponding to the car's steer: $\text{torque} \in [-1.0, 1.0]$, gas: $\text{acceleration} \in [0,0, 1.0]$, and brake: $\text{deceleration} \in [0.0, 1.0]$. Each action is repeated 8 times. The agent receive a $96\times 96 \times 3$ RGB pixel observation. The top $84 \times 96$ portion of the frame contains a clipped, egocentric, bird's eye view of the horizontally centered car. The bottom $12 \times 96$ segment simulates a dashboard visualizing the agent's latest action and return. Snapshots of the test track in the F1 benchmark are shown in Figure~\ref{app:fig-cr-test-images}.

\begin{figure}
    \centering
    \includegraphics[width=\textwidth]{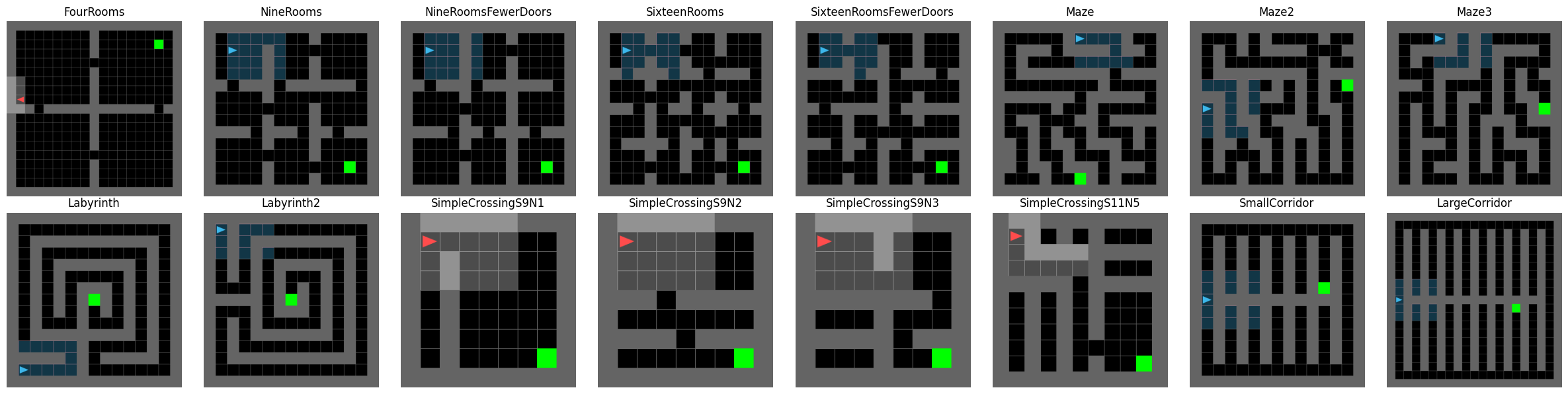}
    \caption{Snapshots of the test grids for MiniGrid}
    \label{app:fig-mg50-test-images}
\end{figure}

\textbf{Minigrid}: The environment is partially observable and based on \cite{minigrid} and adopted for UED by \cite{paired}. Each navigation task is represented with a sequence of integers denoting the locations of the obstacles, the goal, and the starting position of the agent: on a $15\times15$ grid similar to~\cite{paired}. The grids are surrounded by walls on the sides, making it essentially a $13\times13$ grid. \cite{paired} parameterizes the locations using integers. Each task is a sequence of 52 integers, while the first 50 numbers denote the location of obstacles followed by the goal and the agent's initial location. The sequences  may contain duplicates to allow the generation of navigation tasks with fewer than 50 obstacles. Snapshots of the test grids used in our paper are shown in Figure~\ref{app:fig-mg50-test-images}.

\subsection{Network Architectures}
\label{app:network-architecture}

All the student and teacher agents are trained with PPO~\cite{ppo-paper}.

\textbf{Student Architecture}

For CarRacing, we use the same student architecture as~\cite{dcd}. The architecture consists an image embedding module composed of 2D Convolutions with square kernels of sizes 2,2,2,2,3,3, stride lengths 2,2,2,2,1,1 and channel outputs of 8, 16, 64, 128, 256 stacked together. The image embedding is of size 256 and is passed through a Fully Connected (FC) layer of 100 hidden units and then passed through ReLU activations. This embedding is then passed through two FC with 100 hidden neurons, and then a softplus layer, and finally added to 1 for the beta distribution  used for the continuous action space. Further details can be found in ~\cite{dcd}. 

For navigation tasks, we use the same student architecture as ~\cite{paired}. The observation is a tuple with a $5 \times 5 \times 3$ grid observation and a direction integer in $[0-3]$. The grid view is fed to a convolutional layer with kernels of size 3 with 16 filters and the direction integer is passed through a FC with 5 units. This is followed by an LSTM of size 256, and then to two FC layers with 32 units, which connect to the policy outputs. The value network uses the same architecture.

\textbf{Teacher Architecture} 

For CarRacing, CLUTR teacher takes a random noise and generates a continuous vector, i.e., the latent task vector. We pass the random noise through a feed-forward network with one hidden layer of 8 neurons as the teacher. The output of this layer is fed through two separate fully-connected layers, each with a hidden size of 8 and an output dimension equal to the latent space dimension, followed by soft plus activations.  We then add 1 to each component of these two output vectors, which serve as the $\alpha$ and $\beta$ parameters respectively for the Beta distributions used to sample each latent dimension. In all of our experiments, we used a 64-dimensional latent task space.

For Minigrid experiments with flexible regret objective, we use a similar architecture as CarRacing described above, except the  hidden layer consists of 10 neurons, instead of eight. For Minigrid experiments with standard regret objective (which is discussed later in Section~\ref{app:clutrp-mg50}), we use the network architecture used in ~\cite{paired} but only take a noise input. As this adversary network generates discrete actions, we scale them to real numbers before feeding into the VAE decoder.

\textbf{VAE architecture}

We use the architecture proposed in ~\cite{clutr-vae}. We use a word-embedding layer of size 300 with random initialization. The encoder comprises a conditional `Highway' network followed by an LSTM. The Highway network is a two-staged network stacked on top of each other. Each stage computes $\sigma(x) \odot f(G(x)) + (1 - \sigma(x)) \odot Q(x)$, where $x$ is the inputs to each of the highway network stages, G and Q is affine transformation, $\sigma(x)$ is a sigmoid non-linearization, and $\odot$ is element-wise multiplication. $G$ and $Q$ are feed-forward networks with a single hidden layer with equal input and output dimensions of 300, equal to the word-embedding output dimension. We use ReLU activation as $f$. The highway network is followed by a bidirectional LSTM with a single layer of 600 units. The LSTM outputs are passed through linear layer of dimension 64 to get the VAE mean and log variance. The mean vectors are passed through a hyperbolic tangent activation. For CarRacing (both Flexible and Standard Objective experiments) and navigation (only Standard Objective) tasks the output of the hyperbolic tangent activation is linearly scaled in $[-4,4]$. No such scaling is done for the MiniGrid experiments with Flexible Regret Objective. The decoder takes in latent vectors of dimension 64 and passes through a bidirectional LSTM with two hidden layers of size 800 and follows it by a linear layer with size equaling the parameter vector dimension. 

\subsection{Hyperparameters}
All our agents are trained with PPO~\citet{ppo-paper}. We did not perform any hyperparameter search for our experiments. The CarRacing experiments used the same parameters used in~\citet{dcd} and the Minigrid experiments used the parameters from ~\citet{paired}. The VAE used for CarRacing and Minigrid standard objective experiments (Section~\ref{app:clutrp-mg50}) were trained using the default parameters  from ~\citet{clutr-vae}. For the VAE used in the Minigrid flexible objective experiments, which we presented in the main text of the paper, we used a reconstruction weight of 1000 and ran the training for 10M steps to incorporate the larger dataset. The detailed parameters are listed in Table~\ref{app:tab-vae-parameter} and Table~\ref{app:tab-clutr-paired-parameter}.

The flexible objective blurs the distinction between the agent and the antagonist. Hence, we designate the agent achieving the higher average training return during the last 10 steps as the primary student agent and the other one as antagonist.

\begin{table}[!ht]
\begin{tabular}{|l|r|}
\hline
\multicolumn{1}{|c|}{Parameter} & \multicolumn{1}{c|}{Value}             \\ \hline
Batch Size                    & 32                                       \\ 
Number of Training Steps      & 1000000                                  \\ 
Reconstruction Weight         & 79                                       \\ 
Latent Variable Size          & 64                                       \\ 
Word Embedding size           & 300                                      \\ 
Maximum Sequence Length       & 52                                       \\ 
Encoder Activation            & \multicolumn{1}{l|}{Hyperbolic Tangent}  \\ 
Learning Rate                 & 0.00005                                  \\ 
Dropout                       & 0.3                                      \\ \hline
\end{tabular}
\caption{Hyperparameters for training the Task VAE}
\label{app:tab-vae-parameter}
\end{table}

\begin{table}[!ht]
\begin{tabular}{|l|r|r|}
\hline 
Parameter                   & CarRacing               & MiniGrid                \\ \hline
$\gamma$                    & 0.99                    & 0.995                   \\
$\lambda_{GAE}$             & 0.9                     & 0.95                    \\
PPO rollout length          & 125                     & 256                     \\
PPO epochs                  & 8                       & 5                       \\
PPO minibatches per epoch   & 4                       & 1                       \\
PPO clip range              & 0.2                     & 0.2                     \\
PPO number of workers       & 16                      & 32                      \\
Adam learning rate          & 3e-4                & 1e-4                \\
Adam $\epsilon$             & 1e-5                & 1e-5                \\
PPO max gradient norm       & 0.5                     & 0.5                     \\
PPO value clipping          & \multicolumn{1}{r|}{no}  & \multicolumn{1}{r|}{yes} \\
Return normalization        & \multicolumn{1}{r|}{yes} & \multicolumn{1}{r|}{no}  \\
Value loss coefficient      & 0.5                     & 0.5                     \\
Student entropy coefficient & 0                       & 0                       \\
Action Repeat               & 8                       & \multicolumn{1}{r|}{-}   \\ \hline
\end{tabular}
\caption{Hyperparameters for PAIRED and CLUTR PPO training.}
\label{app:tab-clutr-paired-parameter}
\end{table}

\subsection{VAE Training Data}\label{app:vae-training}
For CarRacing, we follow the same parameterization as~\citet{dcd}: each track is defined with a sequence of up to 12 integers denoting control points of a  B\'{e}zier Curve. . Each control point is represented with an integer. We generate 1M random sorted integer sequences of fixed length 12 with duplicates---which enables generating tracks defined with less than 12 control points. 

For navigation tasks we use the parameterization of ~\citet{paired}, generating upto 50 obstacles for each task for a $15 \times 15$ grid, surrounded by walls, effectively an active area of $13\times13$. Hence, each location is numbered in 1 to 169. Every number except the last two of the sequence represent obstacle locations, and the last two for the goal and agent location, respectively.  The parameter vector is thus  partially permutation invariant. We uniformly generate 1M and 10M sequences of variable length between 2 and 52 (inclusive), for the standard regret objective and flexible regret objective, respectively. The obstacle locations are sorted. 

\subsection{Details on Compute Resources}
We have conducted our experiments in cloud machines from :Amazon EC2 - Secure Cloud Services (\url{https://aws.amazon.com/}) and Google Cloud Platform (GCP) - Google Cloud (\url{https://cloud.google.com/}). We used a single NVIDIA T4 GPUs for our experiments with machines having 8(16) and 16(32) physical(virtual) cores, 64GB and 128 GB Memory for CarRacing and Minigrid experiments. A typical 500M Minigrid training of CLUTR ran with a speed of around 800-900 environment interactions per second, taking around 6-8 days, with 32 parallel workers. CarRacing experiments ran on around 90-110 environment interactions per second with 16 parallel processes.

\section{Detailed Experimental results on CarRacing}\label{app:carracing-exp-detailed}

\subsection{Detailed Comparison on Full F1 dataset}\label{app:cr-detailed-comparison}

\begin{figure}[!htbp]
     \centering
     \includegraphics[width=0.48\textwidth]{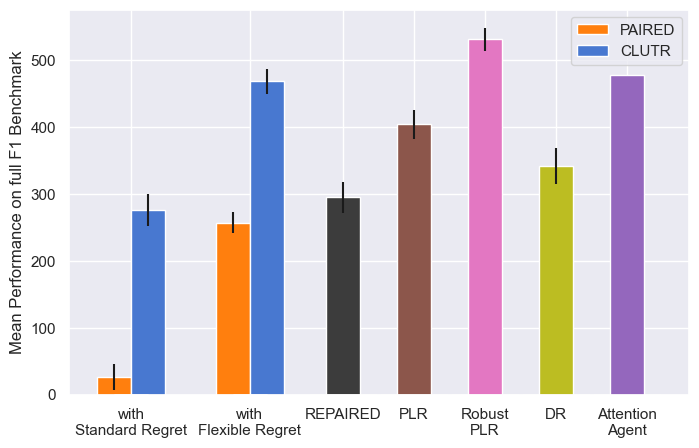}
     \caption{Comparison on the F1 Benchmark comprising 20 tracks modeled on real-life F1 racing tracks. CLUTR (with flexible regret) emerges as the best adaptive-teacher UED for CarRacing and being the only adaptive-teacher UED to outperform some of the random-generator UEDs. Each of the other adaptive-teacher UEDs (REPAIRED, PAIRED with flexible regret, CLUTR with standard regret) are outperformed by all of the random-generator UEDs (DR, PLR, Robust PLR). CLUTR outperforms the adaptive-teacher PAIRED and REPAIRED by 82\% and 58\%, respectively, while outperforming Domain Randomization and PLR, by 38\% and 16\%, repectively. It only falls short to Robust PLR by 14\%. The results show mean and standard error of 10 independent runs.}
    \label{fig:cr-final-comp-all}
\end{figure}

Figure~\ref{fig:cr-final-comp-all} and Table~\ref{tab:clutr-cr-full}, compares CLUTR with contemporary random-generator UED methods, REPAIRED, and the attention based SOTA. It is to be noted that, CLUTR and PAIRED with flexible regret objective was trained for 2M timesteps. All the other UED methods, along with CLUTR and PAIRED with standard regret was trained for 5M timesteps.  

We notice that, each of the random-teacher UEDs outperform each of the adaptive-teacher UEDs, except CLUTR with flexible regret objective, indicating that adaptive-teacher UEDs face significant difficulty in this domain. PAIRED performs miserably in its basic form. CLUTR (with standard regret), flexible regret objecctive, or REPAIRED (by introducing replay  and stop-gradient capabilities), all can improve PAIRED (with standard objective loss) significantly, yet they stil fall short to any of the random-teacher UEDs. 

CLUTR with flexible regret emerges as the best adaptive-teacher UED and the only adaptive-teacher UED to show better performance than some of the random-teacher UEDs-despite being trained only for 2M timesteps. CLUTR with flexible regret achieves an impressive 18X higher zero-shot generalization than PAIRED with standard regret and outperforms REPAIRED by 58\%.

CLUTR with flexible regret is the only adaptive-teacher UED to outperform other random-teacher UEDs. CLUTR outperforms Domain Randomization and PLR, by  38\% and 16\%, repectively. It only falls short to Robust PLR by 14\%. Nonetheless, CLUTR shows competitive results compared to Robust PLR, showing comparable results in seven out of the 20 test tracks and outperforming in the Netherlands track.  CLUTR also outperforms the non-UED SOTA on the full F1 dataset. CLUTR outperforms the Attention Agent on nine out of the 20 tracks and shows comparable performance in another one.

\begin{sidewaystable}[]
\centering
\begin{tabular}{|l|lll|l|ll|ll|r|}
\hline
\multicolumn{1}{|c|}{Track} & \multicolumn{1}{c}{DR} & \multicolumn{1}{c}{PLR} & \multicolumn{1}{c|}{Robust PLR} & \multicolumn{1}{c|}{REPAIRED} & \multicolumn{1}{c}{PAIRED} & \multicolumn{1}{c|}{CLUTR} & \multicolumn{1}{c}{PAIRED} & \multicolumn{1}{c|}{CLUTR} & \multicolumn{1}{c|}{Attention} \\ \cline{6-9}
\multicolumn{1}{|c|}{}      & \multicolumn{1}{c}{}   & \multicolumn{1}{c}{}    & \multicolumn{1}{c|}{}           & \multicolumn{1}{c}{}         & \multicolumn{2}{|c|}{Standard Regret}                    & \multicolumn{2}{|c|}{Flexible Regret (2M)}               & \multicolumn{1}{|c|}{Agent}                \\
\hline
Australia                 & 484 ± 29               & 545 ± 23                & \textbf{692 ± 15}              & 414 ± 27                     & 100 ± 22                   & 429 ± 28                  & 342 ± 29                   & \textbf{683 ± 20}         & 826                                 \\
Austria                   & 409 ± 21               & 442 ± 18                & \textbf{615 ± 13}              & 345 ± 19                     & 92 ± 24                    & 309 ± 19                  & 316 ± 23                   & 507 ± 19                  & \textit{511}                        \\
Bahrain                   & 298 ± 27               & 411 ± 22                & \textbf{590 ± 15}              & 295 ± 23                     & -35 ± 19                   & 225 ± 24                  & 183 ± 28                   & 414 ± 20                  & \textit{372}                        \\
Belgium                   & 328 ± 16               & 327 ± 15                & \textbf{474 ± 12}              & 293 ± 19                     & 72 ± 20                    & 315 ± 14                  & 309 ± 17                   & 429 ± 15                  & 668                                 \\
Brazil                    & 309 ± 23               & 387 ± 17                & \textbf{455 ± 13}              & 256 ± 19                     & 76 ± 18                    & 244 ± 16                  & 237 ± 16                   & 363 ± 18                  & \textit{145}                        \\
China                     & 115 ± 24               & 84 ± 20                 & \textbf{228 ± 24}              & 7 ± 18                       & -101 ± 9                   & 33 ± 19                   & 23 ± 21                    & \textbf{254 ± 28}         & 344                                 \\
France                    & 279 ± 32               & 290 ± 35                & \textbf{478 ± 22}              & 240 ± 29                     & -81 ± 13                   & 266 ± 30                  & 158 ± 24                   & \textbf{498 ± 31}         & \textit{153}                        \\
Germany                   & 274 ± 23               & 388 ± 20                & \textbf{499 ± 18}              & 272 ± 22                     & -33 ± 16                   & 195 ± 26                  & 286 ± 26                   & 404 ± 20                  & \textit{214}                        \\
Hungary                   & 465 ± 32               & 533 ± 26                & \textbf{708 ± 17}              & 414 ± 29                     & 98 ± 29                    & 325 ± 32                  & 327 ± 31                   & 630 ± 24                  & 769                                 \\
Italy                     & 461 ± 27               & 588 ± 20                & \textbf{625 ± 12}              & 371 ± 25                     & 132 ± 24                   & 439 ± 31                  & 451 ± 27                   & \textbf{639 ± 16}         & 798                                 \\
Malaysia                  & 236 ± 25               & 283 ± 20                & \textbf{400 ± 18}              & 200 ± 17                     & -26 ± 17                   & 174 ± 23                  & 192 ± 21                   & \textbf{426 ± 22}         & \textit{300}                        \\
Mexico                    & 458 ± 33               & 561 ± 21                & \textbf{712 ± 12}              & 415 ± 30                     & 67 ± 31                    & 387 ± 31                  & 391 ± 30                   & 627 ± 19                  & \textit{580}                        \\
Monaco                    & 268 ± 28               & 360 ± 32                & \textbf{486 ± 19}              & 256 ± 26                     & -28 ± 18                   & 234 ± 30                  & 125 ± 28                   & \textbf{460 ± 29}         & 835                                 \\
Netherlands               & 328 ± 26               & 418 ± 21                & 419 ± 25                       & 307 ± 21                     & 70 ± 20                    & 302 ± 27                  & 306 ± 24                   & \textbf{488 ± 21}         & \textit{131}                        \\
Portugal                  & 324 ± 27               & 407 ± 15                & \textbf{483 ± 13}              & 265 ± 21                     & -49 ± 13                   & 299 ± 24                  & 149 ± 19                   & \textbf{462 ± 20}         & 606                                 \\
Russia                    & 382 ± 30               & 479 ± 24                & \textbf{649 ± 14}              & 419 ± 25                     & 51 ± 21                    & 319 ± 25                  & 337 ± 24                   & 497 ± 23                  & 732                                 \\
Singapore                 & 336 ± 29               & 386 ± 22                & \textbf{566 ± 15}              & 274 ± 21                     & -35 ± 14                   & 229 ± 18                  & 192 ± 21                   & 382 ± 19                  & \textit{276}                        \\
Spain                     & 433 ± 24               & 482 ± 17                & \textbf{622 ± 14}              & 358 ± 24                     & 134 ± 24                   & 373 ± 15                  & 414 ± 19                   & 496 ± 15                  & 759                                 \\
UK                        & 393 ± 28               & 456 ± 16                & \textbf{538 ± 17}              & 380 ± 22                     & 138 ± 25                   & 396 ± 18                  & 339 ± 18                   & 471 ± 19                  & 729                                 \\
USA                       & 263 ± 31               & 243 ± 28                & \textbf{381 ± 33}              & 120 ± 25                     & -119 ± 11                  & 27 ± 29                   & 67 ± 29                    & 238 ± 31                  & \textit{-192}                       \\
\hline 
Mean                      & 342 ± 27               & 404 ± 22                & \textbf{531 ± 17}              & 295 ± 23                     & 26 ± 19                    & 276 ± 24                  & 257 ± 16                   & 468 ± 21                  & 478 \\                               \hline 
\end{tabular}

\caption{Comparison between CLUTR and other UED algorithms on the individual tracks of the F1 benchmark. We report CLUTR and PAIRED for both standard and flexible regret objectives. We note that, CLUTR and PAIRED with flexible regret was trained for 2M timesteps. All the other UEDs were run for 5M timesteps. Boldface denotes SOTA among UED algorithms, while italic in the Attention Agent column means, CLUTR with Flexible Regret, our best performing model, is comparable/outperforms the attention agent on that track. CLUTR outperforms PAIRED, Domain Randomization, PLR, and REPAIRED and only falls short to Robust PLR. Nonetheless, CLUTR shows comparable results cwith respect to Robust PLR in seven out of the 20 test tracks and outperforming it in the Netherlands track. CLUTR also outperforms the non-UED SOTA on 9 out of the 20 tracks and shows comparableperformance in one.}
\label{tab:clutr-cr-full}
\end{sidewaystable}

\begin{figure}
 \centering
     \includegraphics[width=\textwidth]{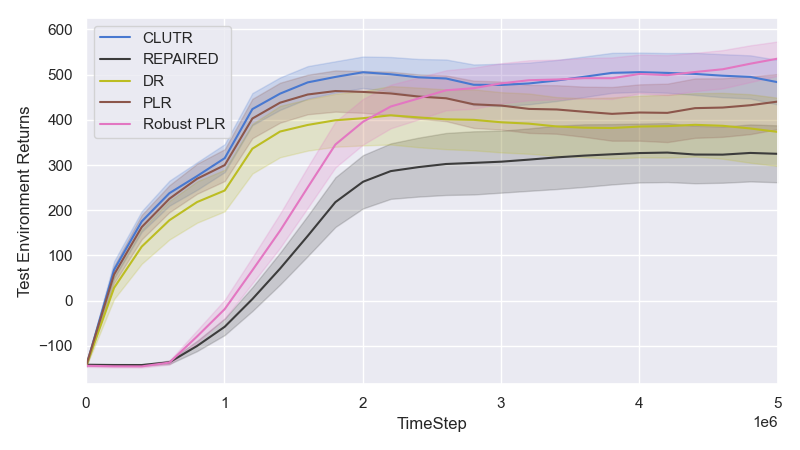}
     \caption{Comparison of mean agent returns on three tracks: Singapore, Germany, and Italy. Based on this subset of tracks, CLUTR (with flexible regret) shows better generalization than all the other UEDs, except Robust PLR. CLUTR was ahead of Robust PLR till around 3M timesteps, followed by both curves following each other closely, and near the very end Robust PLR surpassed CLUTR.} 
     \label{fig:cr-mean-test-all-5M}
\end{figure}

Figure~\ref{fig:cr-mean-test-all-5M} compares how different UEDs perform during training by periodically evaluating them on three tracks from the F1 benchmark: Singapore, Germany, and Italy. CLUTR (with flexible regret) shows better generalization and sample efficiency than all the other UEDs, except Robust PLR. CLUTR showed better performance than Robust PLR till alomost 3M timesteps, after that CLUTR and Robust PLR curves followed each other closely, and near the very end Robust PLR surpasses CLUTR.

\newpage
\subsection{CLUTR with flexible regret loss}
\label{app:CLUTR-CR-Flexible-Regret}

\subsubsection{Training Returns}
\begin{figure}
 \centering
     \includegraphics[width=0.45\textwidth]{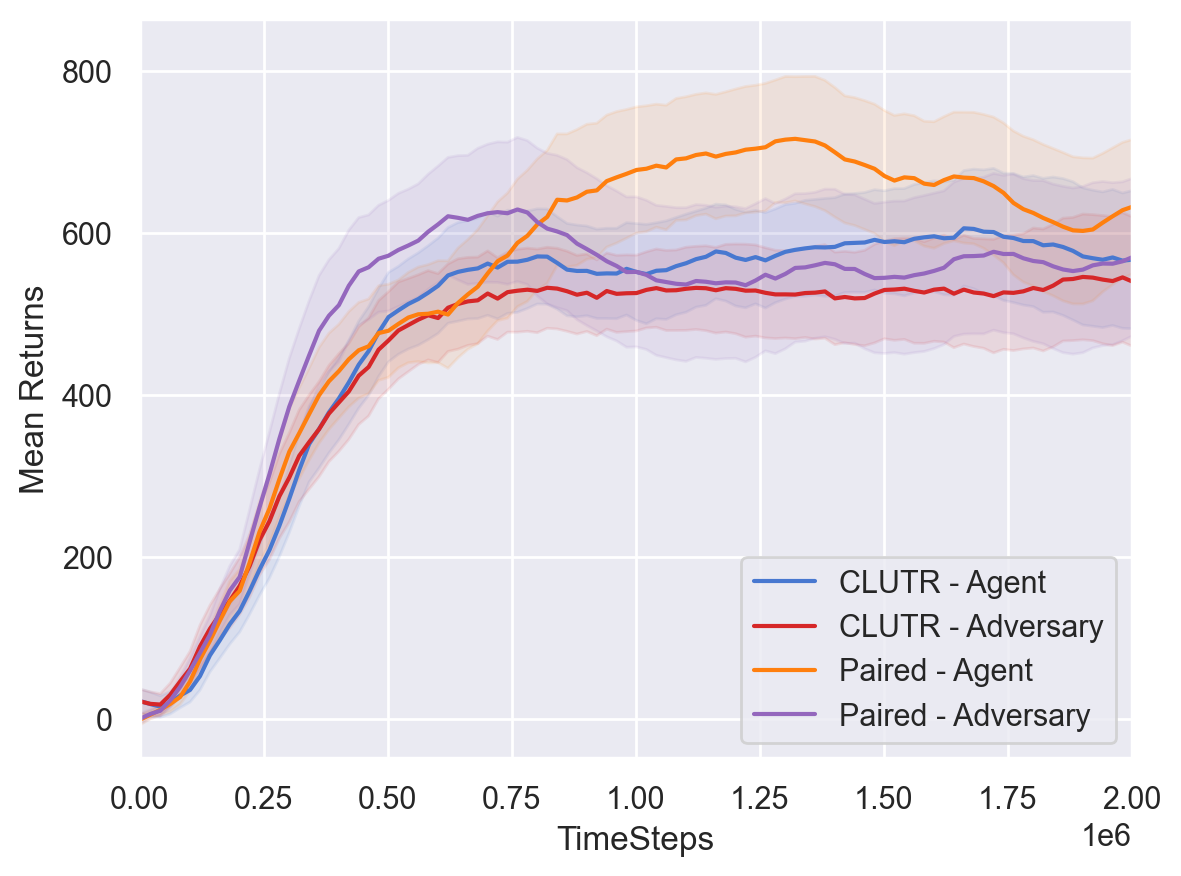}
     \caption{Mean return on the training tasks for both the student agents. CLUTR student agents show close performance, while PAIRED students show a bigger gap of performance between them. Closely competing agents can indicate the training tasks being slightly harder than the agents can currently solve, resulting in a smoother curriculum} 
     \label{fig:cr-flx-train-returns}
\end{figure}

Figure~\ref{fig:cr-flx-train-returns} plot mean return on the training tasks for both the student agents. CLUTR student agents show close performance, while PAIRED students show a bigger gap of performance between them. Closely competing agents can indicate the training tasks being slightly harder than the agents can currently solve.

\subsubsection{Learning task manifold and curriculum: Joint vs
Two-staged Optimization}
\label{app:finetune-vae-cr-flexible}

\begin{figure}[!htbp]
     \centering
     \vspace{-10pt}
     \includegraphics[width=0.48\textwidth]{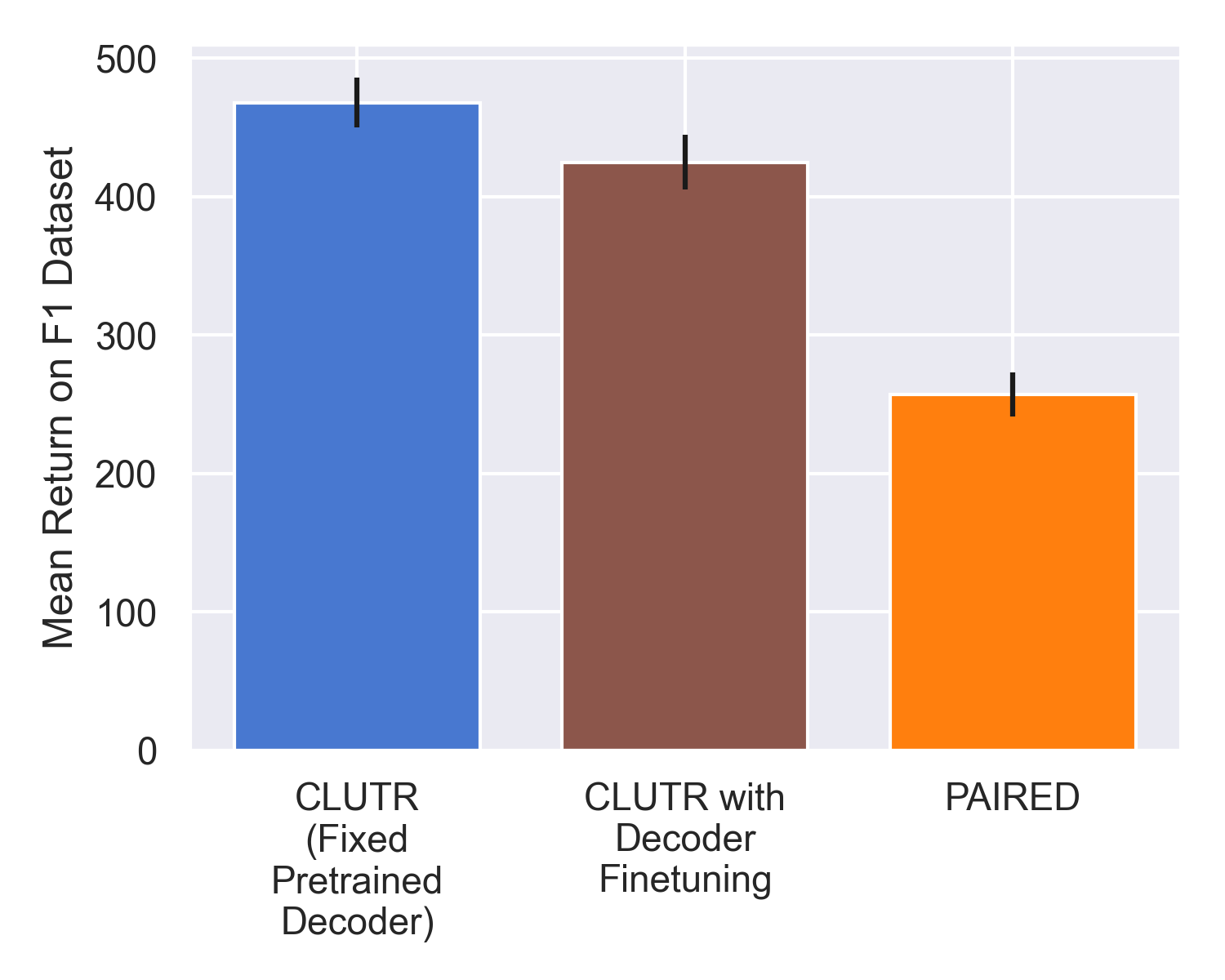}
     \vspace{-15pt}
     \caption{Impact of joint vs two-staged optimization of the task manifold. The leftmost column shows the default CLUTR performance---i.e., using a pretrained decoder kept fixed during the curriculum learning phase---with flexible regret objective in the CarRacing domain. Decoder finetuning, i.e., when the decoder is allowed to finetune with the regret loss,  results in a 10\% performance drop. This performance drop empricially justify our choice of using a pretrained and fixed VAE to solve learning instability. }
    \label{fig:mean-test-clutrfp-vs-clutrft}
\end{figure} 

In Section~\ref{sec:exp-finetune-vae}, we empirically justified our hypothesis that learning the task representation and the curriculum simultaneously results in a difficult learning problem due to the non-stationarity of the process---using the standard regret objective. In this section we repeat the experiment with the flexible regret objective. In Figure~\ref{fig:mean-test-clutrfp-vs-clutrft}, we see a 10\% drop in the performance when the decoder was allowed to finetune with regret loss, further justifying our hypothesis. As a side note, the smaller drop compared to standard regret objective indicates that flexible objective mitigates some of the instability problem too. Finally, even with decoder finetuning, CLUTR achieves a 65\% improvement over PAIRED indicating the benefits of pretrained decoupled latent task space. 

\subsection{CLUTR with standard regret loss}
\label{app:CLUTR-CR-Standard-Regret}

\begin{figure}[!h]
     \centering
     \begin{subfigure}[b]{0.49\textwidth}
         \centering
         \includegraphics[width=\textwidth]{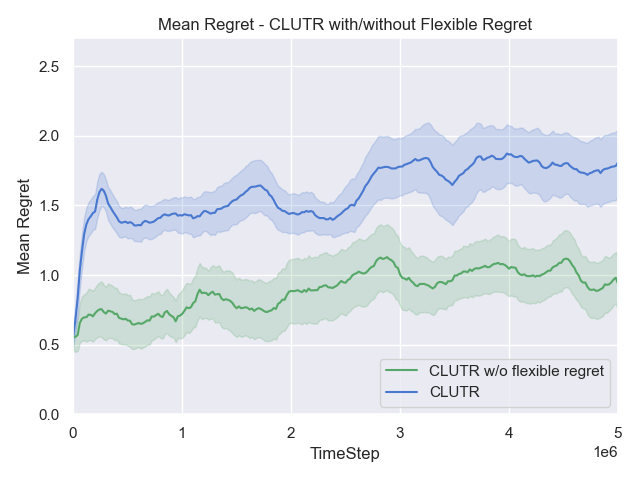}
         \caption{Mean Regret - Car Racing - with vs without flexible regret loss}
         \label{fig:regret-cr-clutrp-vs-clutrfp}
     \end{subfigure}
     \begin{subfigure}[b]{0.49\textwidth}
         \centering
         \includegraphics[width=\textwidth]{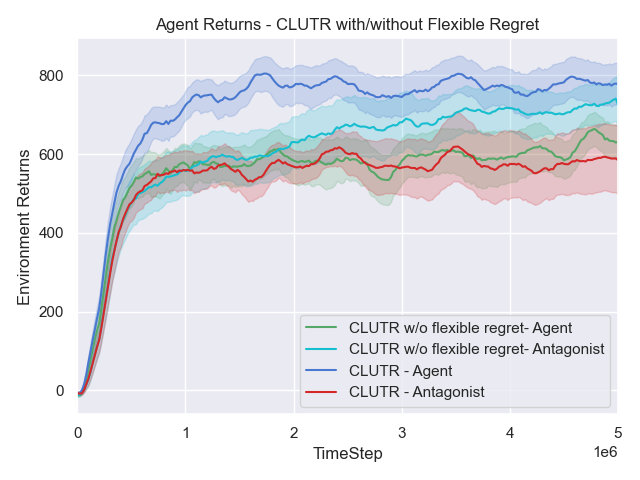}
         \caption{Returns on UED generated Car Racing tracks - with vs without flexible regret loss}
         \label{fig:returns-training-cr-clutrp-vs-clutrfp}
     \end{subfigure}
    \caption{Mean Regret and agent returns during training CLUTR (with flexible regret) vs CLUTR with standard PAIRED regret approximation.}
    \label{fig:clutrp-vs-clutrfp}
\end{figure}

We train CLUTR with the standard regret loss for 5M timesteps. Figure~\ref{fig:clutrp-vs-clutrfp} compares the impact of standard/flexible regret loss on the regret and agent returns during training. With standard regret loss, CLUTR shows a lower regret value, but shows similar pattern. The CLUTR agent achieves better returns with flexible loss throughout the training. 

\begin{figure}
     \centering
     \begin{subfigure}[b]{0.49\textwidth}
         \centering
         \includegraphics[width=\textwidth]{sections/figscarracing/regret_clutrp_vs_paired.png}
         \caption{Mean Regret - Car Racing}
         \label{fig:regret-cr}
     \end{subfigure}
     \begin{subfigure}[b]{0.49\textwidth}
         \centering
         \includegraphics[width=\textwidth]{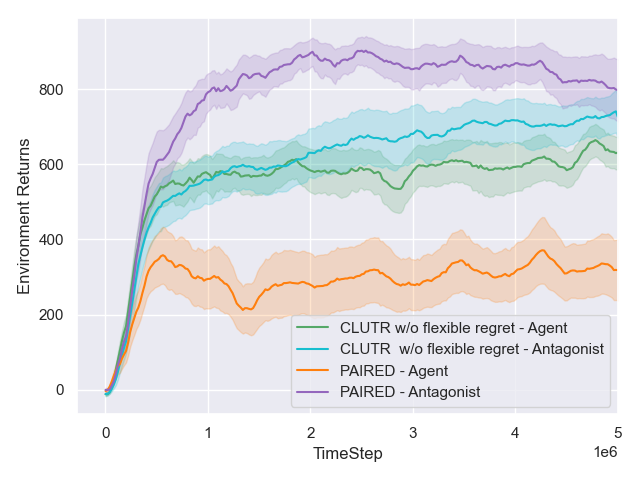}
         \caption{Returns on UED generated Car Racing tracks}
         \label{fig:returns-training-cr}
     \end{subfigure}
    \caption{Mean Regret and agent returns during training CLUTR with standard PAIRED regret loss (i.e., without the flexible regret). CLUTR shows a smaller regret value(i.e., closely competing agent and antagonist), indicating a better UED curriculum. }
    \label{fig:regret}
\end{figure}

\begin{figure}
 \centering
     \includegraphics[width=\textwidth]{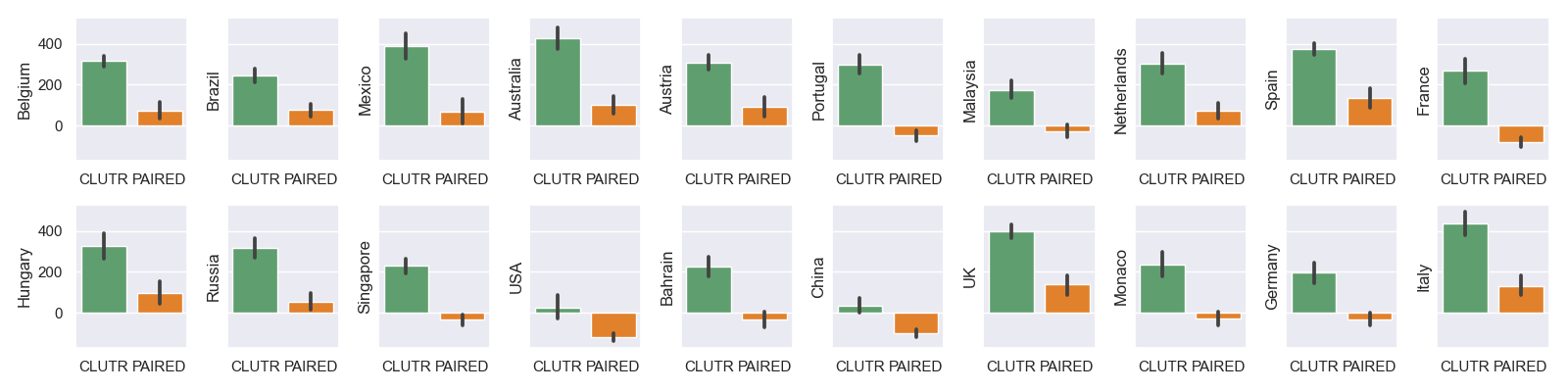}
     \caption{Zero-shot generalization of both PAIRED and CLUTR (with the standard regret loss) agents after 5M timesteps on the full F1 benchmark. CLUTR with the standard regret loss outperforms PAIRED on every track. For each track, we test the agents on 10 different episodes and the error bar denotes the standard error.} 
     \label{fig:all-F1-pairedp-vs-clutr}
\end{figure}

\begin{figure}
     \centering
     \includegraphics[width=0.5\textwidth]{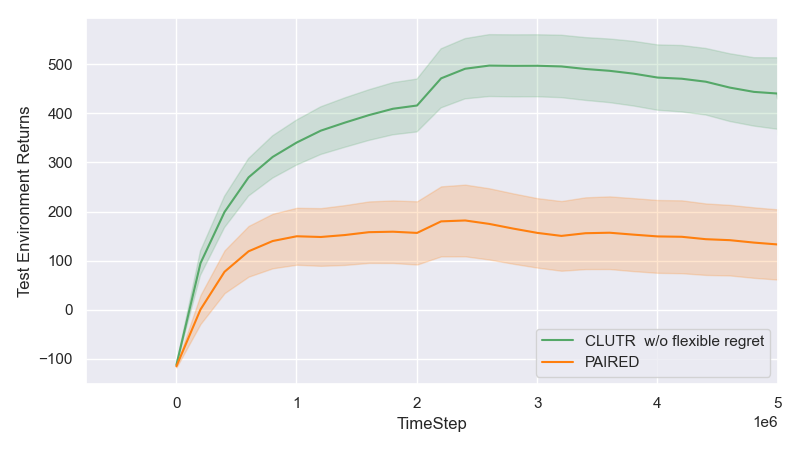}
     \caption{Test Returns on Selected Tracks (Vanilla, Singapore, Germany, and Italy) of CLUTR with standard PAIRED regret loss alongside PAIRED performance.}
     \label{fig:eval_during_training_cr_clutrp}    
\end{figure}

Figure~\ref{fig:regret} compares the mean regret and agent training returns with PAIRED. CLUTR with standard loss shows much lower regret than PAIRED (Figure~\ref{fig:regret-cr}). Figure~\ref{fig:returns-training-cr} shows that the CLUTR agents compete closely, while PAIRED antagonist achieves much higher returns than the PAIRED agent which leads to higher regret returns for the teacher agent but results in a weak student agent. To test the Zero-shot generalization, we evaluate CLUTR with the standard loss on the full F1 benchmark. Figure~\ref{fig:all-F1-pairedp-vs-clutr} shows CLUTR with standard regret loss outperforms PAIRED in all the 20 test tracks. This implies  that CLUTR outperforms PAIRED irrespective of the choice of the loss function (standard/flexible). Figure~\ref{fig:eval_during_training_cr_clutrp} compares the sample efficiency of CLUTR with the standard regret loss with PAIRED by evaluating the agents on four selected tracks (Vanilla, Singapore, Germany, Italy) during training. It can be seen that CLUTR, even without the regret loss, outperforms PAIRED significantly. We note that these test environments were not used in any way, neither during training CLUTR (and PAIRED) nor while designing it. 

As mentioned in~\cite{dcd} PAIRED overexploits the relative strengths of the antagonist over the protagonist and generates a curriculum that gradually reduces the task complexity. However, CLUTR overcomes this and generates a curriculum where the agent and the antagonist closely compete (Figure~\ref{fig:returns-training-cr}) and shows a robust generalization on the unseen F1 benchmark. 

\subsection{Extended Analysis on Impact of sorting training data for VAE training}
~\label{app:shuffled-vae}

\begin{figure}[!h]
     \centering
     \begin{subfigure}[b]{0.49\textwidth}
         \centering
         \includegraphics[width=\textwidth]{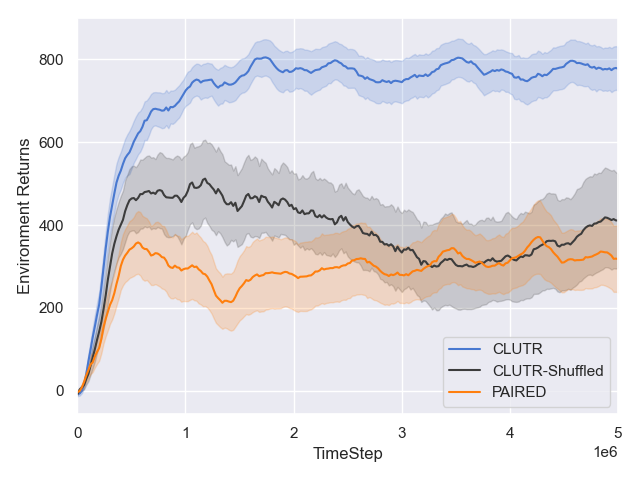}
         \caption{During training CLUTR agent achieves higher returns while, CLUTR-shuffled agent shows lower returns. CLUTR-Shuffled agent's  return is also less stable showing a decrease and increase.}
         \label{fig:agent-returns-clutrfp-vs-clutr-shuffled}
     \end{subfigure}
     \begin{subfigure}[b]{0.49\textwidth}
         \centering
         \includegraphics[width=\textwidth]{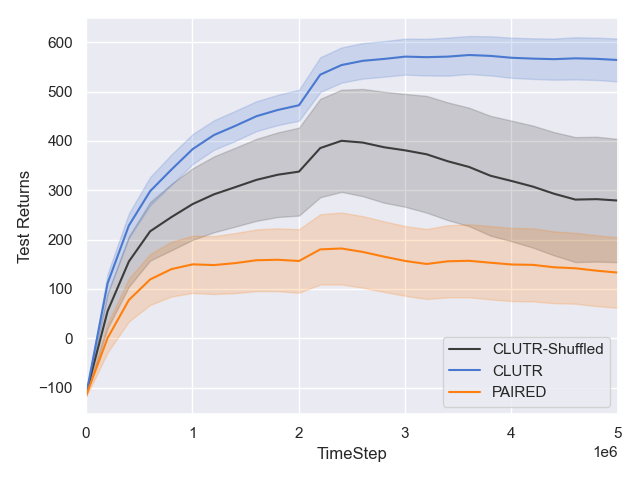}
         \caption{CLUTR achieves higher and more stable mean returns on the selected tracks. CLUTR-Shuffle shows signs of unlearning.}
         \label{fig:returns-training-cr-clutrfp-vs-clutr-shuffled}
     \end{subfigure}
    \caption{Analysis of sorting training data for VAE. Trained on shuffled data, CLUTR-Shuffled performs inferior compared to CLUTR and shows signs of unlearning.}
    \label{fig:clutrfp-vs-clutrsh}
\end{figure}

The non-sorted dataset was generated by shuffling each track of the original VAE training dataset 10 different times, resulting in a 10X bigger dataset (10M tracks). It was trained for 5X longer for 5M training steps. We planned on training for 10M gradient steps (10X than the original VAE) but stopped at 5M as it converged much sooner.   We ran both CLUTR and CLUTR-shuffled, i.e., CLUTR with a VAE trained on non-sorted data up to 5M timesteps. CLUTR-shuffled shows inferior performance and also signs of unlearning compared to CLUTR.  Figure~\ref{fig:clutrfp-vs-clutrsh} shows detailed experiment results.

\subsection{Impact of Task Representation Learning}
\begin{wrapfigure}[19]{r}{0.5\textwidth}
     \centering
     \vspace{-15pt}
     \includegraphics[width=0.5\textwidth]{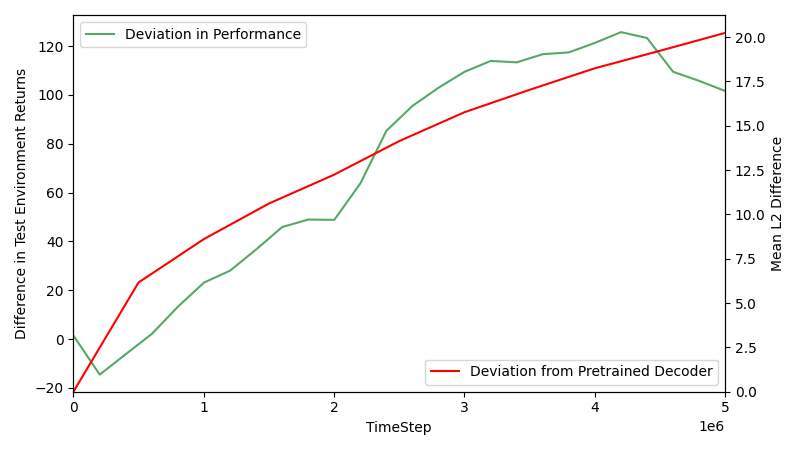}
     \caption{Impact of pretrained decoder weights on performance. The red curve plots the deviation of the decoder from its pretrained weights as it is finetuned. The green curve shows the performance drop from CLUTR with the standard loss. These curves suggest that pretrained weights are crucial for performance.}
    \label{fig:finetune-deviation}
\end{wrapfigure}
In this section, we discuss the impact of the learned task representation on performance. In Section~\ref{sec:exp-finetune-vae}, we showed that if we finetune the VAE decoder during curriculum learning, the overall performance drops significantly (Figure~\ref{fig:mean-test-clutr-vs-clutrsh-vs-clutrft}). To get a better understanding, in Figure~\ref{fig:finetune-deviation}, we plot how much the performance deviates as the VAE decoder changes during the training process. The curve in red shows the deviation of the decoder from its pretrained weights as it is fine-tuned during the training. We estimate the deviation as the L2 distance between the  finetuned and the pretrained decoder weights. The green curve shows the performance drop from CLUTR (with standard loss). To estimate the performance drop, we periodically evaluate both CLUTR and CLUTR with Finetuned VAE, on the selected test tracks during training. From the figure, we observe that, as the decoder weights are finetuned, they become increasingly different from the initial pretrained weights. At the same time, the overall performance gap from CLUTR also increases.  This suggests that the pretrained VAE weights are crucial for better performance. 

Furthermore, the quality of the learned representation depends on the quality of the data they are trained on. In section~\ref{sec:exp-sorting-vae-data}, we showed that a VAE trained on a non-sorted dataset significantly deteriorates the performance (Figure~\ref{fig:mean-test-clutr-vs-clutrsh-vs-clutrft}). This further suggests that the learned representation has a significant impact on performance. We also want to note that both of these variations (CLUTR with Finetuned VAE and the CLUTR with Shuffled VAE) perform much better than PAIRED, which suggests that, though CLUTR's performance depends on the representation, with a reasonable representation, it can still perform better than PAIRED.


\section{Detailed Experimental results on  on MiniGrid}
\label{app:mg-details}

\subsection{CLUTR with flexible regret objective}
\label{app:clutrfp-mg50}

To train the CLUTR VAE, we generated 10 million random grids, with the obstacle locations sorted, and the number of obstacles  uniformly varying from zero to 50, aligning with~\cite{paired}. We train both CLUTR and PAIRED using the flexible regret objectives.

\begin{figure*}[!hbtp]
 \centering
     \includegraphics[width=\textwidth]{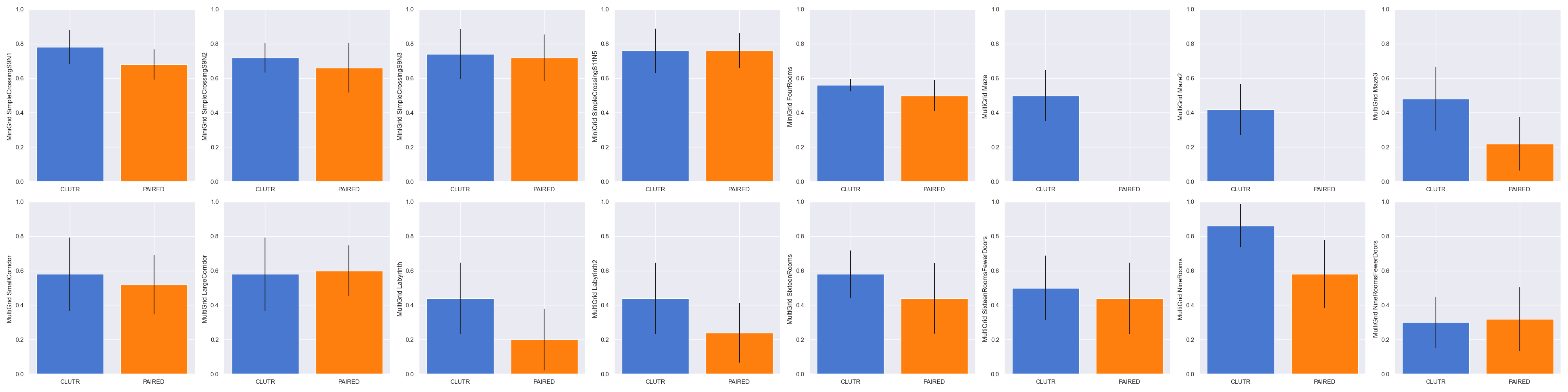}
     \caption{Zero-shot generalization of CLUTR and PAIRED, in terms of percent of the environments solved. CLUTR achieves a higher solved rate than PAIRED in 13 out of the 16 tasks. We evaluate the agents with 10 independent episodes on each task. Error bars denote the standard error.} 
     \label{fig:all-mg-pairedfp-vs-clutrfp}
\end{figure*}

Figure~\ref{fig:all-mg-pairedfp-vs-clutrfp} shows zero-shot generalization performance of CLUTR and PAIRED on the 16 unseen navigation tasks from~\citet{paired}, in terms of the percent of environments the agent solved, i.e., solved rate. CLUTR achieves a 1.35X better generalization solving 58\% of the unseen grids, than PAIRED which solves 43\% of the unseen grids. It can also be seen that CLUTR outperforms PAIRED on 13 out of the 16 test navigation tasks. 

\begin{figure}
 \centering
     \includegraphics[width=0.45\textwidth]{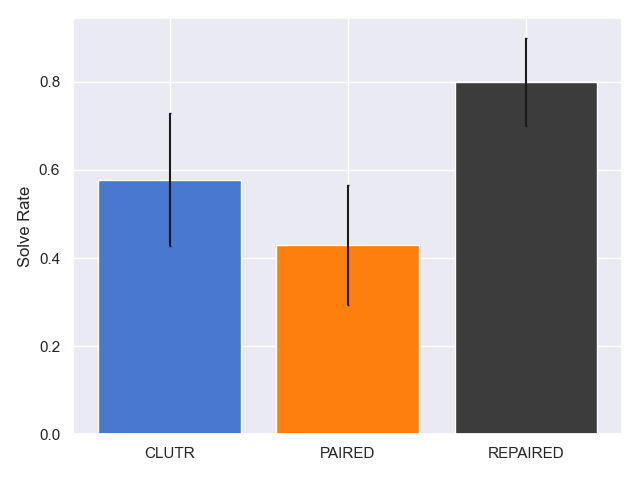}
     \caption{Mean solve rate on Minigrid testset. REPAIRED outperforms both CLUTR and PAIRED.} 
     \label{fig:mg-flx-compare-repaired}
\end{figure}

Figure~\ref{fig:mg-flx-compare-repaired} compared the mean perforamnce of CLUTR, PAIRED, and REPAIRED. REPAIRED outperforms both PAIRED and CLUTR. We note that, REPAIRED and CLUTR are both improvement towards PAIRED. However, REPAIRED involves a dual-curriculum methods, with two different teachers adopting replay capabilities with disabling exploratory gradients. On the other hand CLUTR is a much simpler method, and can also be augmented with REPAIRED too.  

\subsubsection{Training Returns}
\begin{figure}
 \centering
     \includegraphics[width=0.45\textwidth]{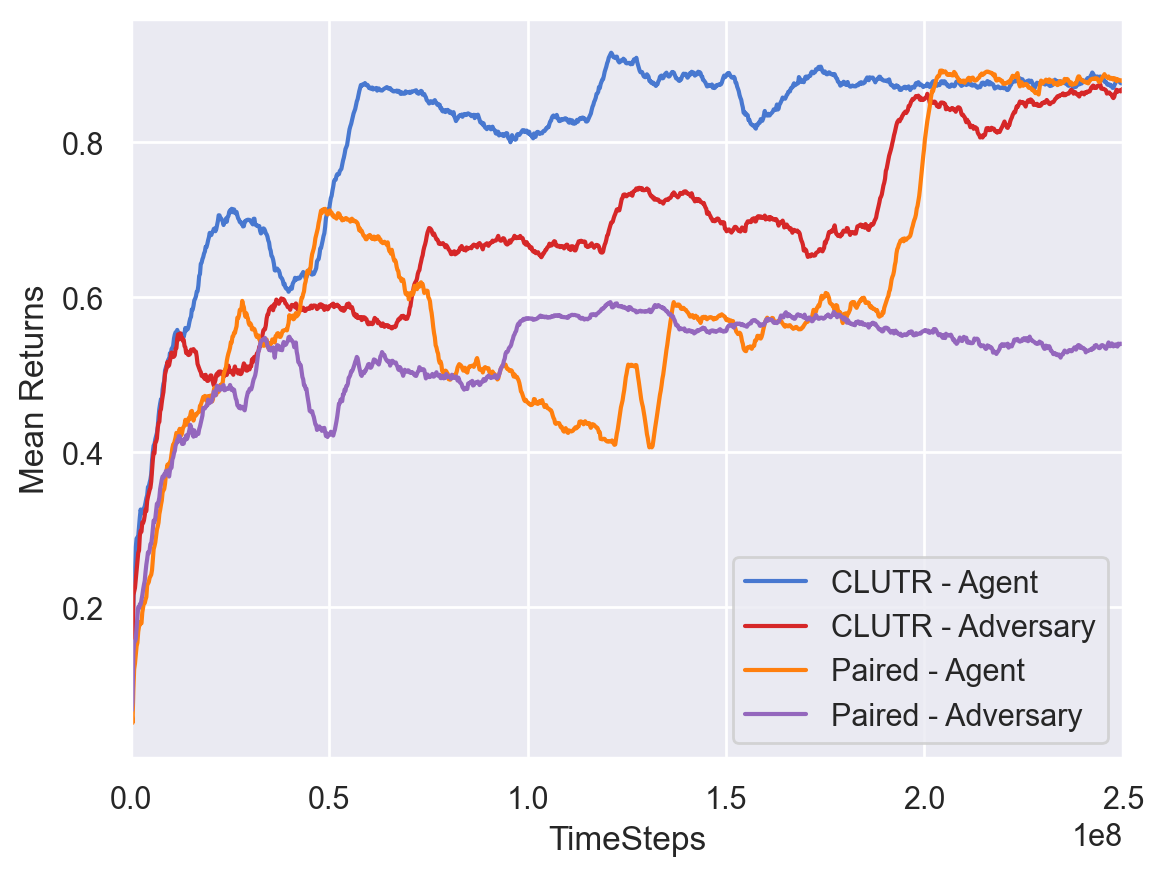}
     \caption{Mean return on the training tasks for both the student agents. CLUTR student agents show close performance, while PAIRED students show a bigger gap of performance between them. Closely competing agents can indicate the training tasks being slightly harder than the agents can currently solve, resulting in a smoother curriculum} 
     \label{fig:mg-flx-train-returns}
\end{figure}

Figure~\ref{fig:mg-flx-train-returns} plot mean return on the training tasks for both the student agents. CLUTR student agents show close performance, while PAIRED students show a bigger gap of performance between them. 

\subsection{CLUTR with standard regret objective}
\label{app:clutrp-mg50}

\subsubsection{Training Returns}
\begin{figure}
 \centering
     \includegraphics[width=0.45\textwidth]{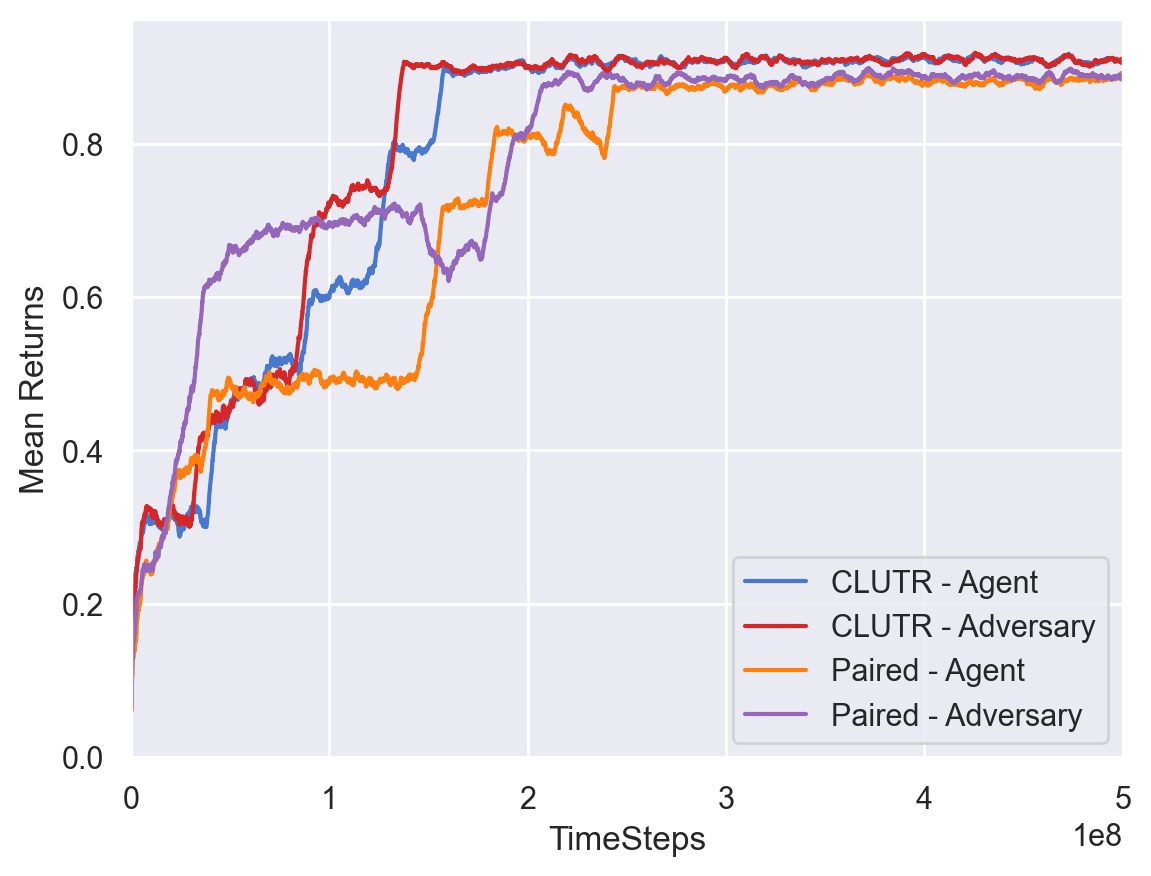}
     \caption{Mean return on the training tasks for both the student agents. CLUTR student agents show close performance, while PAIRED students show a bigger gap of performance between them initiallly at the beginning. } 
     \label{fig:mg-std-train-returns}
\end{figure}

Figure~\ref{fig:mg-std-train-returns} plot mean return on the training tasks for both the student agents. CLUTR student agents show close performance, while PAIRED students show a bigger gap of performance between them initiallly at the beginning. 

\subsection*{Performance}
\begin{figure}[!htbp]
     \centering
     \includegraphics[width=0.48\textwidth]{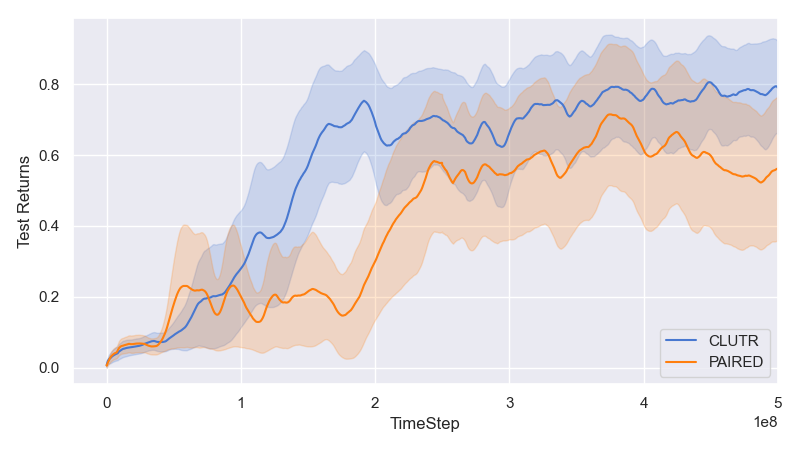}
     \caption{Agent solved rate on selected grids during training. CLUTR shows better sample efficiency and generalization than PAIRED. The results show an average of 5 independent runs.}
    \label{fig:eval_during_training_mg50_standard}
\end{figure}

\begin{figure*}[!hbtp]
 \centering
     \includegraphics[width=\textwidth]{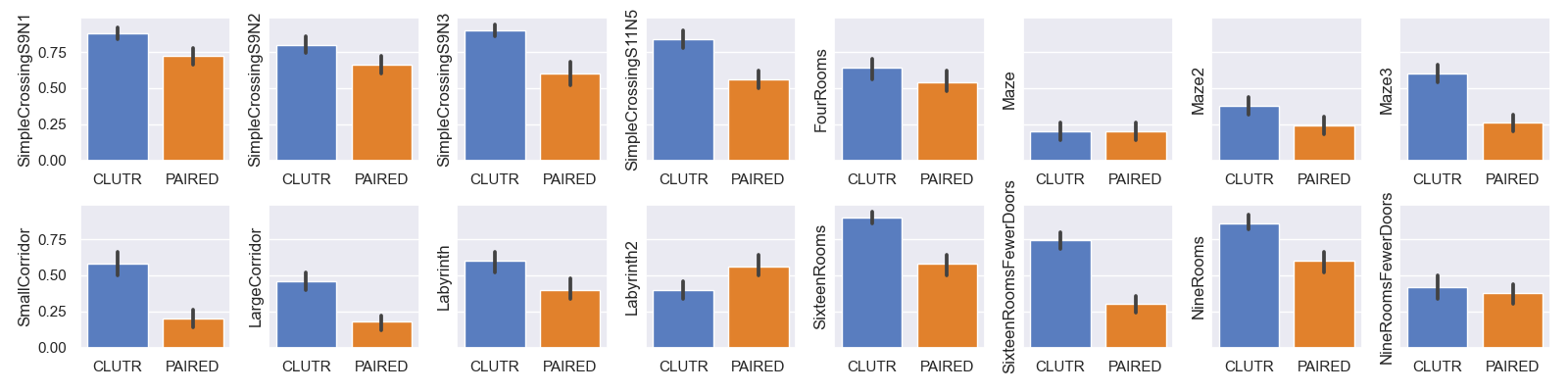}
     \caption{Zero-shot generalization of CLUTR and PAIRED, in terms of percent of the 14  solved. CLUTR achieves a higher solved rate than PAIRED in 14 out of the 16 unseen tasks. We evaluate the agents with 100 independent episodes on each task. Error bars denote the standard error.} 
     \label{fig:all-mg-pairedp-vs-clutrp}
\end{figure*}

Figure~\ref{fig:all-mg-pairedp-vs-clutrp} shows zero-shot generalization performance of CLUTR and PAIRED on 16 unseen navigation tasks from~\cite{paired} based on the percent of environments the agent solved, i.e., solved rate. CLUTR achieves superior generalization solving 64\% of the unseen grids, a 45.45\% improvement over PAIRED, which achieves a 44\% solve rate. From figure~\ref{fig:all-mg-pairedp-vs-clutrp} it can be seen CLUTR outperforms PAIRED achieving a higher mean solve rate on 14 out of the 16 unseen navigation tasks. Figure~\ref{fig:eval_during_training_mg50_standard} shows solved rates on four selected grids (Sixteen Rooms, Sixteen Rooms with Fewer Doors, Labyrinth, and Large Corridor) during training. CLUTR shows better sample efficiency, as well as generalization than PAIRED.

\subsection*{Curriculum Snapshot}
\begin{figure}[!hptb]
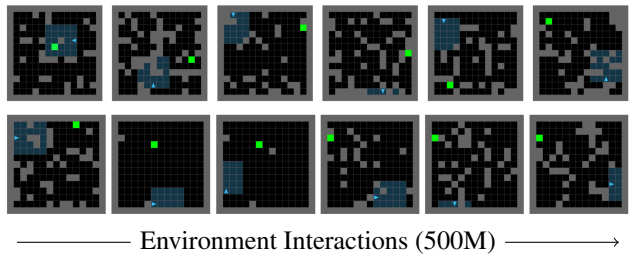

     \centering
     \begin{subfigure}[b]{0.49\textwidth}
         \centering
         \includegraphics[width=\textwidth]{sections/figsmg50/clutr_image_strip.png}
     \end{subfigure}
     \\
     \centering
      \begin{subfigure}[b]{0.49\textwidth}
         \centering
         \includegraphics[width=\textwidth]{sections/figsmg50/paired_image_strip.png}
     \end{subfigure} 
     \\
     \centering
      \begin{subfigure}[b]{0.49\textwidth}
         \centering
            \begin{tikzpicture}
\coordinate (A) at (-4.0,0);
            \coordinate (B) at ( 4.0,0);
            \path (A) -- node (success) {Environment Interactions (500M)} (B);
            \draw[->] (A) -- (success) -- (B);
          \end{tikzpicture}
     \end{subfigure}
     
    \caption{Example grids(right) generated by CLUTR(top) and PAIRED(bottom) uniformly sampled at different stages of training. The training progresses from left to right. }
    \label{fig:curriculum_snapshot_mg50_clutrp_pairedp}
\end{figure}

In this section, we visually inspect the curriculum generated by CLUTR and PAIRED, with snapshots of tasks generated by these methods during different stages of the training (Figure~\ref{fig:curriculum_snapshot_mg50_clutrp_pairedp}). We illustrate one common mode of failure/ineffectiveness shown by PAIRED: The curriculum starts with arbitrarily complex tasks, which none of the agents can solve at the initial stage of training. After a while, PAIRED starts generating rudimentary degenerate tasks. While kept training, PAIRED eventually gets out of the degenerative local minima, and the curriculum complexity starts to emerge. On the other hand, CLUTR does not show such degeneration and generates seemingly interesting tasks throughout. 

\subsection*{Curriculum Analysis}
\label{app:curriculum-analysis-mg}

\subsubsection*{CLUTR vs PAIRED}
Figure~\ref{fig:3d-hist-mg50} shows 3D Histograms showing the frequency of the generated grids against the total number of obstacles they contain. PAIRED starts with a high number of obstacles and then degenerates quickly into grids with very few numbers of obstacles and stays similar for a significant number of steps. Eventually, the number of obstacles increases sharply, converging into a band of around 20 to 40 obstacles on average. On the other hand, in CLUTR, the number of obstacles starts flat, centers around a peak around the middle but still with a wide interval for some number of steps, and the peak drops slightly while the interval stays almost the same. After the `convergence',  PAIRED rarely generates grids with fewer or more obstacles than the band it converges to. On the contrary, CLUTR still generates grids with few or many blocks, which might help to address unlearning or improve the agents on grids with more obstacles, respectively. The above observations illustrate that we can achieve a more efficient curriculum learning without making the problem too easy early or without focusing on a narrow interval with a flat distribution later. Instead, we can start with a wide interval and gradually focus on a peak around the middle without making the interval very narrow. 

\begin{figure}[!h]
     \centering
     \begin{subfigure}[b]{0.49\textwidth}
         \centering
         \includegraphics[width=\textwidth]{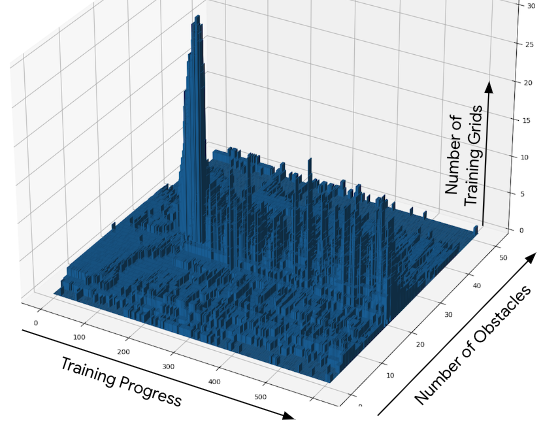}
         \caption{CLUTR}
         \label{fig:3d-hist-clutr-mg50}
     \end{subfigure}
     \begin{subfigure}[b]{0.49\textwidth}
         \centering
         \includegraphics[width=\textwidth]{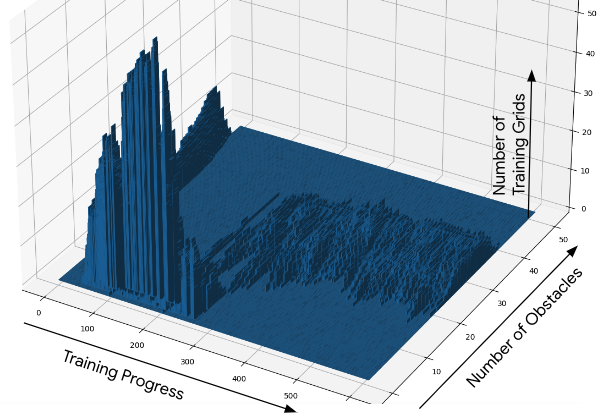}
         \caption{PAIRED}
         \label{fig:3d-hist-paired-mg50}
     \end{subfigure}
    \caption{3D Histograms showing the frequency of the generated grids against the total number of blocks they contain. Both PAIRED and CLUTR converge to a similar band of grids. However, CLUTR converges much faster.}
    \label{fig:3d-hist-mg50}
\end{figure}

\begin{figure}[!h]
     \centering
     \begin{subfigure}[b]{0.49\textwidth}
         \centering
        \includegraphics[width=\textwidth]{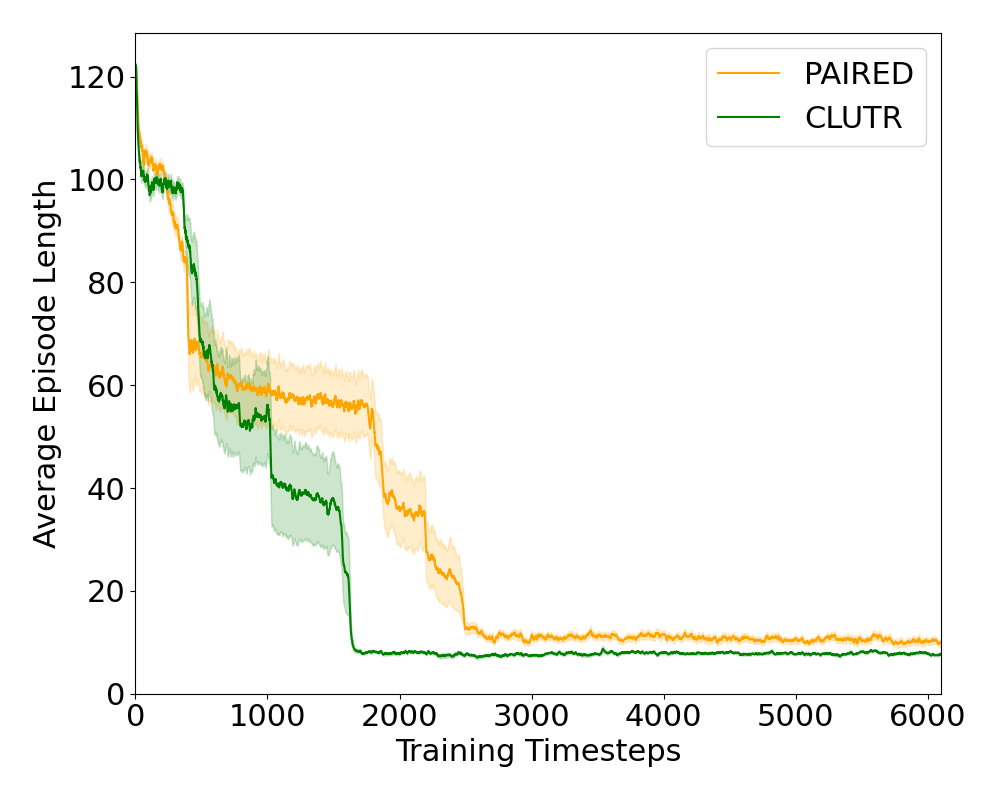}
         \caption{Average length of the training episodes. CLUTR converges sooner than PAIRED to a shorter episode length.}
        \label{fig:mg50-epilen-paired-vs-clutr}
     \end{subfigure}
     \begin{subfigure}[b]{0.49\textwidth}
         \centering
        \includegraphics[width=\textwidth]{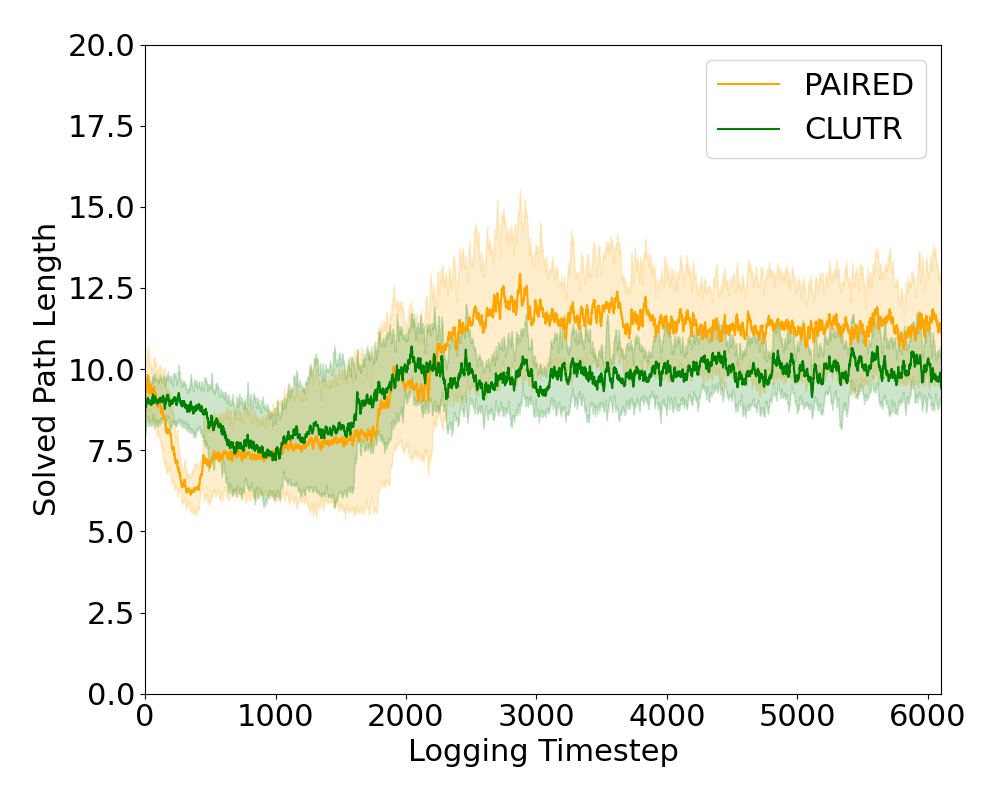}
         \caption{Average solution length of the solved training tasks. }
        \label{fig:mg50-solved-len-paired-vs-clutr}
     \end{subfigure}
    \caption{Comparison of CLUTR and PAIRED curriculum based on properties of the generated grids.}
    \label{fig:3d-hist-mg50}
\end{figure}

Figure~\ref{fig:mg50-epilen-paired-vs-clutr} shows the average episode lengths of both CLUTR and PAIRED. The curves show both methods start with long episodes---indicating at the beginning, the agents do not solve the training grids consistently, and many of the episodes end due to timeout. As the agents learn, the episodes become shorter for both methods until they converge to a small value. However, CLUTR converges sooner than PAIRED.

We also compare the average solution length of the solved training grids. Both PAIRED and CLUTR show a similar pattern. However, PAIRED converges to a larger value than CLUTR. This might indicate that CLUTR is solving the environments more efficiently. This might also mean that CLUTR is solving some easier tasks (e.g., fewer obstacles, as we noticed from Figure~\ref{fig:3d-hist-mg50}) even after convergence lowering its average solved path length slightly.

\subsubsection*{CLUTR curriculum vs. Random Latent Curricula}

\begin{figure}[!h]
     \centering
     \begin{subfigure}[b]{0.32\textwidth}
         \centering
         \includegraphics[width=\textwidth]{sections/figsmg50/3d_hist_blocks_clutr_100_labelled.png}
         \caption{CLUTR}
         \label{fig:3d-hist-clutr-mg50-dr-compare}
     \end{subfigure}
     \begin{subfigure}[b]{0.3\textwidth}
         \centering
         \includegraphics[width=\textwidth]{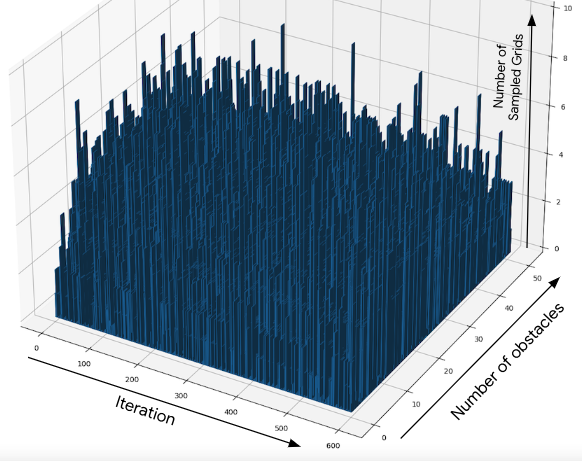}
         \caption{Domain Randomization on the pretrained Latent Space}
         \label{fig:3d-hist-DR-mg50}
     \end{subfigure}
      \begin{subfigure}[b]{0.3\textwidth}
         \centering
         \includegraphics[width=\textwidth]{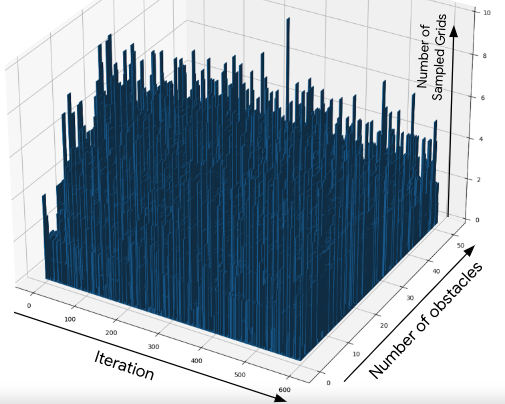}
         \caption{Random Teacher on pretrained Latent Space}
         \label{fig:3d-hist-random-teacher-mg50}
     \end{subfigure}
    \caption{3D Histograms showing the frequency of the CLUTR generated grids against the total number of blocks they contain vs. Domain Randomization on the latent space vs. A random  teacher curriculum on the pretrained latent space. The figures clearly show that CLUTR generates a curriculum significantly different from random curriculums.}
    \label{fig:3d-hist-mg50-compare}
\end{figure}

We further compare CLUTR curriculum with two different domain randomized curriculums. First we compare CLUTR curriculum with a uniform random (i.e., Domain Randomization) curriculum on the latent space by repeatedly sampling the trained VAE (the same VAE used by CLUTR) with a uniform random distribution. Second, we generate a curriculum generated by a random teacher acting on the pretrained latent space. The random teacher uses the same architecture and  intialization procedure as the original CLUTR teacher it is being compared to.  Figure~\ref{fig:3d-hist-mg50-compare} shows the comparison characterizing the grids by the number of obstacles they contain similarly as the previous section. As expected, we can see that the DR and random teacher curriculum generates grids with obstacles ranging from 0 to 50 without showing any pattern, showing  significant difference in the curricula generated by CLUTR and the domain randomized baselines. 

\subsubsection*{Analysis of the Latent Task Manifold}
\begin{wrapfigure}{r}{0.25\textwidth}
 \centering
     \includegraphics[width=0.25\textwidth]{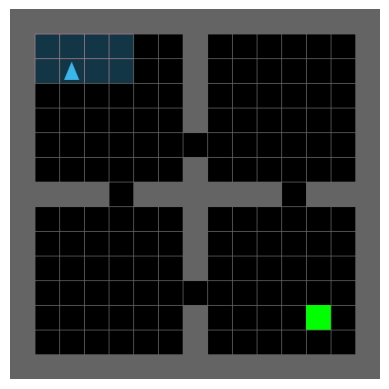}
     \caption{15X15 FourRooms}
    \label{fig:mg-fourroom-15x15}
\end{wrapfigure}
\begin{figure}
     \centering
     \includegraphics[width=\textwidth]{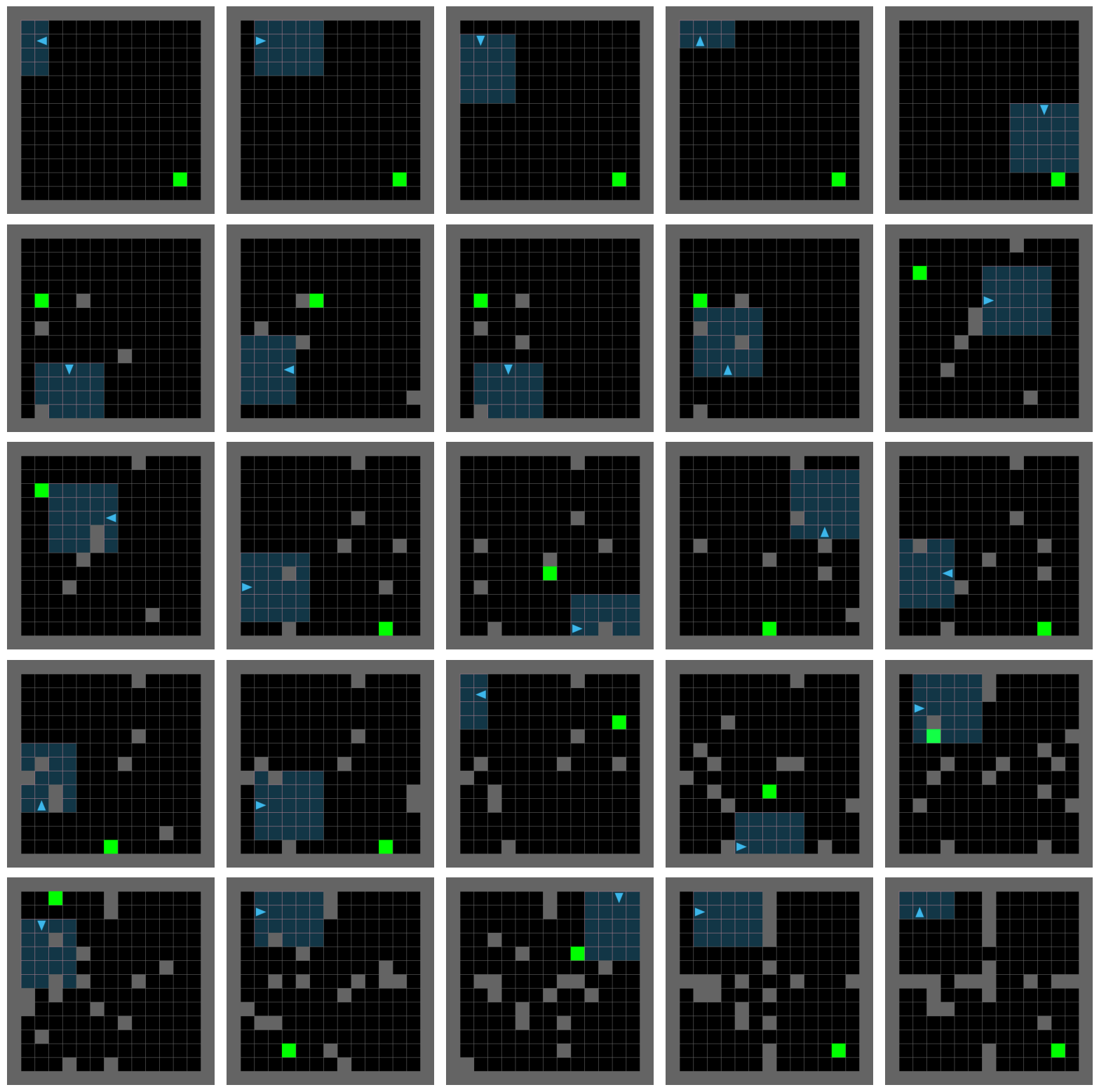}
     \caption{A linear interpolation between an empty grid and 15x15 version of the Four-Room grid (Figure~\ref{fig:mg-fourroom-15x15}) in the latent space. The grids are organized from top-left to bottom-right in row-major order.}
     \label{fig:mg50_interpolation}    
\end{figure}

To grow a sense of the latent task manifold, we linearly interpolate in the latent space between an empty grid and a 15x15 version of the FourRoom grid (shown in Figure~\ref{fig:mg-fourroom-15x15}). Figure~\ref{fig:mg50_interpolation} visualizes the interpolation results. We first get the latent vectors of the empty grid and the target FourRoom task using the VAE encoder. We then linearly interpolate 23 equidistant points between them. At last, we reconstruct the grids from these vectors using our decoder. From Figure~\ref{fig:mg50_interpolation} we see that, as we interpolate in the latent space, the reconstructed grid incrementally adds more obstacles and the grids start to look more like the FourRoom target grid. We note that the reconstruction is not perfect. We also note that the increase in the number of obstacles is not uniform, e.g., the first 5 reconstructed grids are all empty grids, and more obstacles are added near the target point.  Overall, this experiment provides an insight that the latent space holds a useful structure, which CLUTR teacher utilizes to generate the curriculum.

